\pdfoutput=1

\documentclass[11pt]{article}

\usepackage[]{acl}

\usepackage{times}
\usepackage{latexsym}
\usepackage{adjustbox}
\usepackage{amsfonts}
\usepackage{amsmath}
\usepackage{amssymb}
\usepackage{bm}
\usepackage{booktabs}
\usepackage{colortbl}
\usepackage{comment}
\usepackage{dsfont}
\usepackage{enumitem}
\usepackage{graphicx} 
\usepackage{hyperref}
\usepackage{mathtools}
\usepackage{microtype}
\usepackage{multicol}
\usepackage{multirow}
\usepackage{nicefrac}
\usepackage{lscape}
\usepackage{scalerel}
\usepackage{subcaption}
\usepackage{tabularx}
\usepackage[most]{tcolorbox}
\usepackage{xcolor}
\usepackage{xspace}

\newcommand\myfontsize{\fontsize{9.6pt}{10.5pt}\selectfont}

\newcommand\blankfootnote[1]{%
  \let\thefootnote\relax\footnotetext{#1}%
  \let\thefootnote\svthefootnote%
}

\newcommand{\aclcr}[1]{#1}
\newcommand{\ssc}[1]{{\small \sc #1}\xspace}
\newcommand{\bssc}[1]{{\small \sc \textbf{#1}}\xspace}
\newcommand{\fssc}[1]{{\scriptsize \sc #1}\xspace}
\newcommand{\spot}{{\ssc{SPoT}}\xspace}
\newcommand{\bspot}{{\bssc{SPoT}}\xspace}
\newcommand{\bmmc}[1]{\bm{\mathcal{#1}}}
\newcommand{\smallsup}[1]{\scaleto{\text{#1}}{4pt}}

\newcommand{\smallurl}[1]{ \begin{tiny}\url{#1}\end{tiny}}

\definecolor{myblue}{HTML}{2B79B0}
\definecolor{myorange}{HTML}{FB7F36}
\definecolor{mygreen}{HTML}{389E3B}
\definecolor{spotcolor}{HTML}{58A76A}

\usepackage[T1]{fontenc}

\usepackage[utf8]{inputenc}

\usepackage{microtype}

\title{SPoT: Better Frozen Model Adaptation through Soft Prompt Transfer}

\author{Tu Vu$^{1,2}$$^\bigstar$\hspace{8mm}Brian Lester$^{1}$\hspace{8mm}Noah Constant$^{1}$\hspace{8mm}Rami Al-Rfou$^{1}$\hspace{8mm}Daniel Cer$^{1}$\\
  Google Research$^1$\\
  University of Massachusetts Amherst$^2$ \\
  \texttt{\{ttvu,brianlester,nconstant,rmyeid,cer\}@google.com}\\ 
  \texttt{tuvu@cs.umass.edu}\\}

\date{}

\begin{document}
\maketitle

\begin{abstract}
\label{section:abstract}
There has been growing interest in parameter-efficient methods to apply pre-trained language models to downstream tasks.
Building on the \ssc{PromptTuning} approach of \citet{BLester21}, which learns task-specific soft prompts to condition a frozen pre-trained model to perform different tasks, we propose a novel prompt-based transfer learning approach called \spot: \textbf{S}oft \textbf{P}r\textbf{o}mpt \textbf{T}ransfer. %
\spot first learns a prompt on one or more source tasks %
and then uses it to initialize the prompt for a target task.
We show that \spot
significantly boosts the performance of \ssc{PromptTuning} across many tasks. More remarkably, across all model sizes,
\spot matches or outperforms standard \ssc{ModelTuning} (which fine-tunes all model parameters) \aclcr{on the \ssc{SuperGLUE} benchmark}, while using up to 27,000$\times$ fewer task-specific parameters. To understand where \spot is most effective, we conduct a large-scale study on task transferability with 26 \ssc{NLP} tasks in 160 combinations, and demonstrate that many tasks can benefit each other via prompt transfer. Finally, we propose an efficient retrieval approach that interprets task prompts as task embeddings to identify similar tasks and predict the most transferable source tasks for a novel target task.
\blankfootnote{$\bigstar$ Work done during an internship at Google Research.}

\end{abstract}
\section{Introduction}
\label{section:introduction}
The past few years have seen the rapid development of ever larger pre-trained language models, where it has repeatedly been shown that scaling up the model size is a key ingredient for achieving the best performance~\cite{JDevlin19,CRaffel20,TBrown20}.
While this trend has continued to push the boundaries of possibility across various \ssc{NLP} benchmarks, the sheer size of these models presents a challenge for their practical application. For 100B+ parameter models, fine-tuning and deploying a separate instance of the model for each downstream task would be prohibitively expensive.%
\begin{figure}[t!]
\centering
\includegraphics[width=0.48\textwidth]{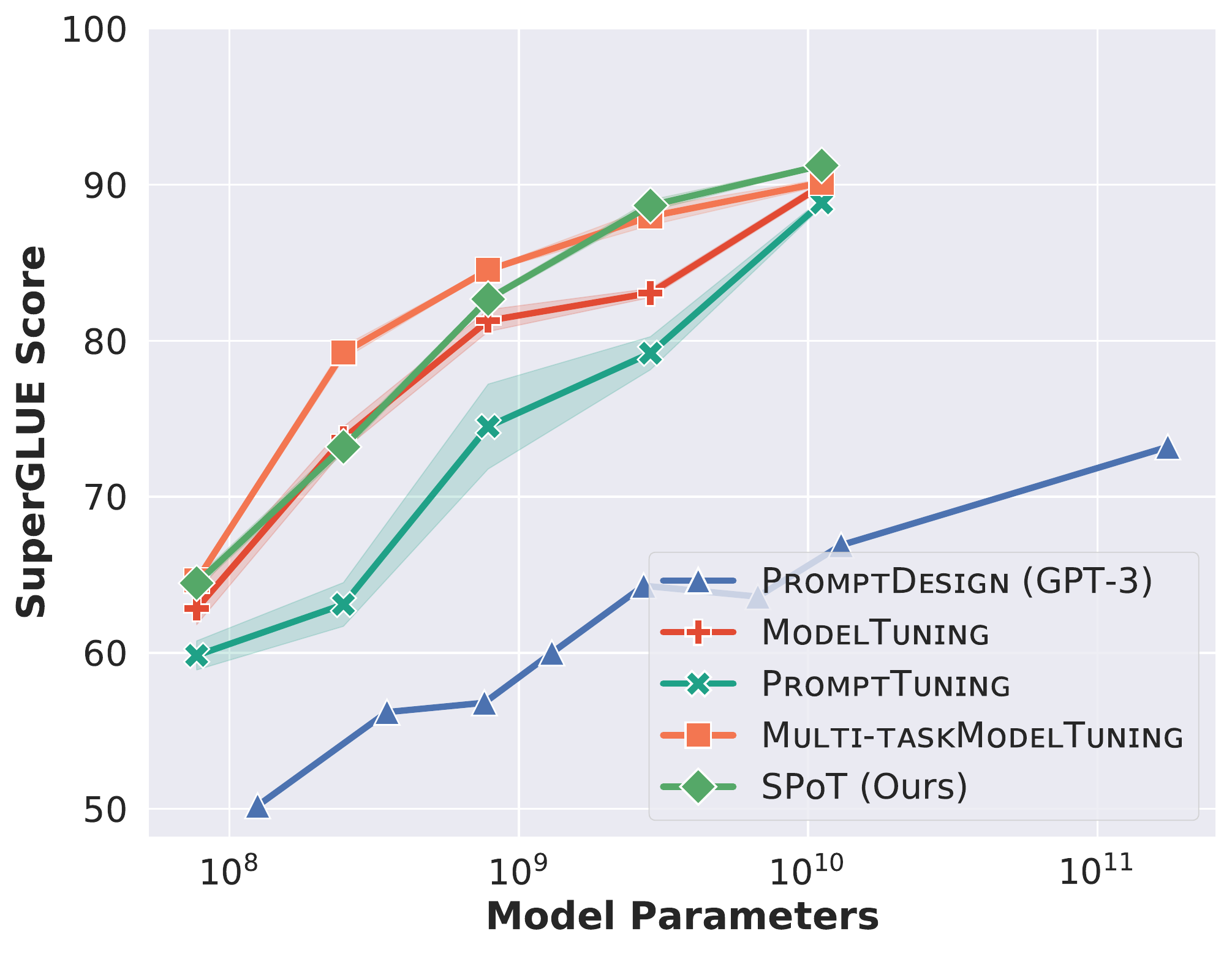}
\caption{Our \spot approach---which transfers a prompt learned from a mixture of source tasks (here, \ssc{GLUE}) onto target tasks---outperforms vanilla %
\ssc{PromtTuning}~\cite{BLester21} and \ssc{\mbox{GPT-3}}~\cite{TBrown20} on \ssc{SuperGLUE} by a large margin, matching or outperforming
\ssc{ModelTuning} across all model sizes. At the \ssc{XXL} model size, \spot even outperforms %
\ssc{Multi-taskModelTuning}, which fine-tunes the entire model on the \ssc{GLUE} mixture before fine-tuning it on individual \ssc{SuperGLUE} tasks. \aclcr{See Appendix~\ref{appendix:fig1_full_results} for full results.}}
\label{fig:superglue_graph}
\vspace{-2mm}
\end{figure}
\begin{figure*}[t!]
\centering
\includegraphics[width=\textwidth]{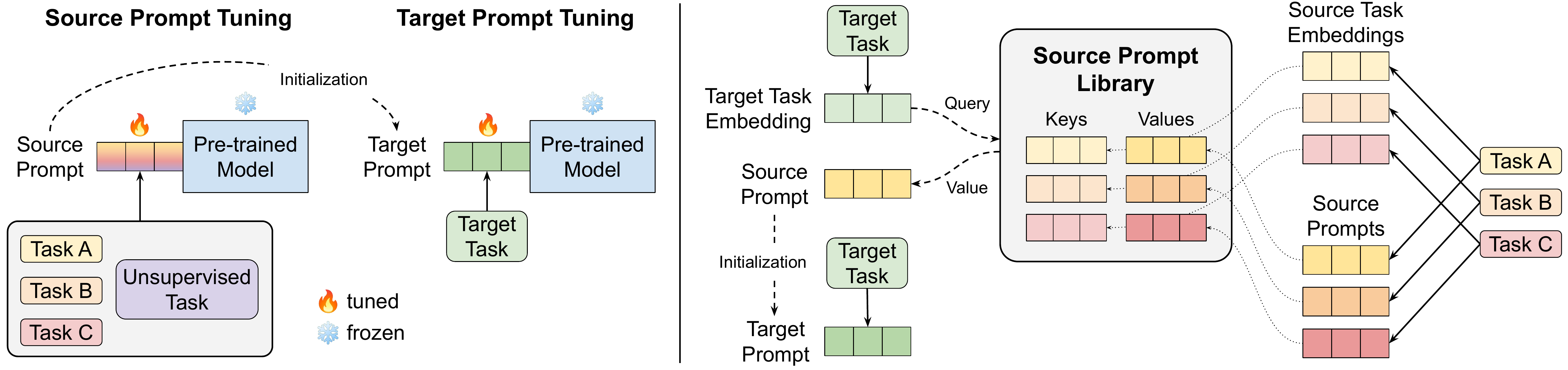}
\caption{An illustration of our \textit{generic} (left) and \textit{targeted} (right) \spot approaches. \textbf{Left}: We learn a single generic source prompt on one or more source tasks, which is then used to initialize the prompt for each target task. \textbf{Right}: We learn separate prompts for various source tasks, saving early checkpoints as task embeddings and best checkpoints as source prompts. These form the keys and values of our prompt library. Given a novel target task, a user: (i)~computes a task embedding, (ii)~retrieves an optimal source prompt, and (iii)~trains a target prompt, initialized from the source prompt (see \S\ref{sec:task-specific_spot} for details).
}
\vspace{-2mm}
\label{fig:approach}
\end{figure*}
To get around the infeasibility of fine-tuning, \citet{TBrown20} propose \ssc{PromptDesign}, where every downstream task is cast as a language modeling task and the \textit{frozen} pre-trained model performs different tasks by conditioning on manual text prompts provided at inference time.
They demonstrate impressive few-shot performance with a single frozen \ssc{\mbox{GPT-3}} model, although its performance depends highly on the choice of the prompt~\cite{TZhao21} and still lags far behind state-of-the-art fine-tuning results.

More recent work has explored methods for learning \textit{soft prompts}~\cite{XLiu21,GQin21,XLi21,BLester21}, which can be seen as additional learnable parameters injected into the language model.
\citet{BLester21} propose \ssc{PromptTuning}, a simple method that learns a small task-specific prompt (a sequence of tunable tokens prepended to each example) for each downstream task during adaptation to condition the frozen language model to perform the task. Strikingly, as model capacity increases, %
\ssc{PromptTuning} %
becomes competitive with %
\ssc{ModelTuning}, %
which fine-tunes the entire model on each downstream task. Nevertheless, at smaller model sizes (below 11B parameters), there are still large gaps between \ssc{PromptTuning} and \ssc{ModelTuning}.%

In this paper, we propose \spot: \textbf{S}oft \textbf{P}r\textbf{o}mpt \textbf{T}ransfer, a novel transfer learning approach in the context of prompt tuning. %
\spot first trains a prompt on one or more source tasks, and then uses the resulting prompt to initialize the prompt for a target (downstream) task. %
Our experiments show that \spot offers significant improvements over \ssc{PromptTuning} across tasks and model sizes. For instance, on the \ssc{SuperGLUE} benchmark~\cite{AWang19b}, we obtain +10.1 and +2.4 point average accuracy improvements using the \ssc{T5 Base} (220M parameter) and \ssc{T5 XXL} (11B parameter) models~\cite{CRaffel20}, respectively. More importantly, \spot is competitive with or outperforms \ssc{ModelTuning} across all model sizes (see Figure~\ref{fig:superglue_graph}).

Motivated by these results, we investigate transferability between tasks, through the lens of soft task prompts. Our goal is to answer two questions: (a)~\textit{For a given target task, when does initializing the prompt from a source task boost performance?} (b)~\textit{Can we use task prompts to efficiently predict which source tasks will transfer well onto a novel target task?}
To answer (a), we conduct a systematic study of the \ssc{T5} model using 26 \ssc{NLP} tasks in 160 combinations of source and target tasks. Our results indicate that many tasks can benefit each other via prompt transfer. %
To address (b), we interpret the learned task prompts as \textit{task embeddings} to construct a semantic space of tasks and formalize the similarity between tasks.
We design an efficient retrieval algorithm that measures task embedding similarity, allowing practitioners to identify source tasks that will likely yield positive transfer. 

To summarize, our main contributions are: (1)~We propose \spot, a novel prompt-based transfer learning approach, and show that scale is not necessary for \ssc{PromptTuning} to match the performance of \ssc{ModelTuning}; \aclcr{on \ssc{SuperGLUE}}, \spot matches or beats \ssc{ModelTuning} across all model sizes. (2)~We conduct a large-scale and systematic study on task transferability, demonstrating conditions under which tasks can benefit each other via prompt transfer. (3)~We propose an efficient retrieval method that interprets task prompts as task embeddings to construct a semantic space of tasks, and measures task embedding similarity to identify which tasks could benefit each other. (4)~To facilitate future work on prompt-based learning, we will release our library of task prompts and pre-trained models, and provide practical recommendations for adapting our library to \ssc{NLP} practitioners at \href{https://github.com/google-research/prompt-tuning/tree/main/prompt_tuning/spot}{\ttfamily\myfontsize\aclcr{\;\;\:https://github.com/google-research/\\ prompt-tuning/tree/main/prompt\_tuning/\\spot}}.

\section{Improving \bssc{PromptTuning} with \bspot}
\label{sec:prompt_pretraining}
To improve performance of \ssc{PromptTuning} \aclcr{on a target task}, \spot introduces \textit{source prompt tuning}, an intermediate training stage between language model pre-training and target prompt tuning (Figure~\ref{fig:approach}, left), to learn a prompt on one or more source tasks (while still keeping the base model frozen), which is then used to initialize the prompt for the target task.\footnote{\aclcr{The target task can be treated as one of the source tasks being mixed together.}} Our approach retains all the computational benefits of \ssc{PromptTuning}: for each target task, it only requires storing a small task-specific prompt, enabling the reuse of a single frozen pre-trained model across all tasks. In this section, we present a \textit{generic} \spot approach where a single transferred prompt is reused for all target tasks. In~\S\ref{sec:task-specific_spot}, we explore a \textit{targeted} approach that retrieves different source prompts for different target tasks.

\subsection{Experimental setup}
Our frozen models are built on top of the pre-trained \ssc{T5} checkpoints of all sizes: \ssc{Small}, \ssc{Base}, \ssc{Large}, \ssc{XL}, \ssc{XXL} with 60M, 220M, 770M,
3B, and 11B parameters, respectively. In our experiments with \spot, we leverage the LM adapted version of \ssc{T5}\footnote{\fssc{T5 1.1} checkpoints trained for an additional 100K steps using the ``prefix LM'' objective~\cite{CRaffel20}, available at \href{https://github.com/google-research/text-to-text-transfer-transformer/blob/main/released_checkpoints.md}{\smallurl{https://github.com/google-research/text-to-text-transfer-transformer/blob/main/released\_checkpoints.md}} 
}, which was found to be easier to optimize for \ssc{PromptTuning}~\cite{BLester21}.
\subsubsection{Baselines}
\label{baselines}
We compare \spot to the following baselines:
\paragraph{\bssc{PromptTuning}:} The vanilla prompt tuning approach of~\citet{BLester21}, where an independent prompt is directly trained on each target task. 

\paragraph{\bssc{ModelTuning \&~Multi-taskModelTuning}:} We compare prompt tuning approaches to \ssc{ModelTuning}, the standard fine-tuning approach~\cite{JDevlin19,CRaffel20}, where all model parameters are fine-tuned on each target task separately. For an apples-to-apples comparison, we include \ssc{Multi-taskModelTuning}, a more competitive baseline that first fine-tunes the entire model on the same mixture of source tasks used for \spot before fine-tuning it on individual target tasks.{\interfootnotelinepenalty=10000\footnote{%
In preliminary experiments, we found that using the original version of \fssc{T5 1.1} (which was pre-trained exclusively on span corruption) for model tuning approaches results in better performance than using the LM adapted version. We therefore report results corresponding to the original \fssc{T5 1.1} for \fssc{ModelTuning} and \fssc{Multi-taskModelTuning}.}}
\subsubsection{Evaluation datasets}
We study downstream performance on a diverse set of tasks from the \ssc{GLUE}~\cite{AWang19a} and \ssc{SuperGLUE}~\cite{AWang19b} benchmarks
.\footnote{These datasets include grammatical acceptability judgments (\fssc{CoLA}~\cite{AWarstadt19}), sentiment analysis (\fssc{SST-2}~\cite{RSocher13}), paraphrasing/semantic similarity (\fssc{MRPC}~\cite{WDolan05}, \fssc{STS-B}~\cite{DCer17}, \fssc{QQP}~\cite{SIyer17}), natural language inference (\fssc{MNLI}~\cite{AWilliams18}, \fssc{QNLI}~\cite{AWang19a}, \fssc{RTE}~\cite[et seq.]{IDagan06}, \fssc{CB}~\cite{MDeMarneffe19}), coreference resolution (\fssc{WSC}~\cite{HLevesque12}), sentence completion (\fssc{COPA}~\cite{MRoemmele11}), word sense disambiguation (\fssc{WiC}~\cite{MPilehvar19}), and question answering (\fssc{MultiRC}~\cite{DKhashabi18}, \fssc{ReCoRD}~\cite{SZhang18}, \fssc{BoolQ}~\cite{CClark19}). We exclude the problematic \fssc{WNLI}~\cite{HLevesque12} dataset from \fssc{GLUE}, following \citet{JDevlin19}.} 
We train for a fixed number of steps and report results on the validation set associated with each dataset.\footnote{For tasks with multiple metrics, we average the metrics.} %

\subsubsection{Data for source prompt tuning}
As with language model pre-training, the choice of training data is crucial for successful prompt transfer. To investigate the impact of source training data on downstream performance, we compare a diverse set of source tasks. %

\paragraph{A single unsupervised learning task:} We first consider training the prompt on a fraction of the \ssc{C4} (Colossal Clean Crawled Corpus) dataset~\cite{CRaffel20} using the ``prefix LM'' objective discussed in~\citet{CRaffel20}. Although this task was used to pre-train our frozen \ssc{T5} models already%
, it could still be helpful for learning a general-purpose prompt.

\paragraph{A single supervised learning task:} Alternatively, we can train the prompt using a supervised task. We use either \ssc{MNLI}~\cite{AWilliams18} or \ssc{SQuAD}~\cite{PRajpurkar16} as a single source task. \ssc{MNLI} was shown to be helpful for many sentence-level classification tasks~\cite{JPhang19}, while \ssc{SQuAD} was found to generalize well to \ssc{QA} tasks~\cite{ATalmor19}.
\paragraph{A multi-task mixture:} %
So far, we have been using a single source task. An alternative approach is multi-task training. Within \ssc{T5}'s unified text-to-text framework, this simply corresponds to mixing different datasets together.
We explore mixing datasets from different \ssc{NLP} benchmarks or families of tasks, including \ssc{GLUE}, \ssc{SuperGLUE}, natural language inference (\ssc{NLI}), paraphrasing/semantic similarity, sentiment analysis, question answering (\ssc{QA}) on \ssc{MRQA}~\cite{AFisch19}, commonsense reasoning on \ssc{RAINBOW}~\cite{NLourie21}, machine translation, summarization, and natural language generation on \ssc{GEM}~\cite{SGehrmann21}.\footnote{See Appendix~\ref{appendix:source_datasets} for details about datasets.} We create a mixture of source tasks from each of the \ssc{NLP} benchmarks/families of tasks above, and a mixture comprising all datasets (C4 + 55 labeled datasets), using the examples-proportional mixing strategy in \citet{CRaffel20} with an artificial dataset size limit $\bmmc{K}=2^{19}$ examples. %
\subsubsection{Training details}
\label{sec:training_details}
We closely follow the training procedure in~\citet{BLester21}. Specifically, the only new parameters introduced during both source and target prompt tuning are a shared prompt $\bm{\rho} \in \mathbb{R}^{\bmmc{L} \times \bmmc{E}}$ prepended to each (embedded) input sequence, where $\bmmc{L}$, $\bmmc{E}$ are the prompt length and the embedding size, respectively. In all cases, we set $\bmmc{L} = 100$ tokens and tune the prompt for a fixed number of steps $\bmmc{S}$.\footnote{We use the Adafactor optimizer~\cite{NShazeer18} with default parameters except with a constant learning rate of 0.3, weight decay of $1e{-5}$, and parameter scaling turned off. We train with a batch size of 32. The dropout probability is always kept at $0.1$. All of our models are implemented using \fssc{JAX}~\cite{JBradbury18} and \fssc{Flax}~\cite{JHeek20}.} While $\bmmc{S}$ is set to 30K in \citet{BLester21}, we find that additional tuning is helpful on large datasets. As such, we set $\bmmc{S}$ to $2^{18} = 262{,}144$, following~\citet{CRaffel20}, with the exception of ablation experiments (rows ``$-$ longer tuning'') in Table~\ref{tbl:pretraining_results} which use $\bmmc{S}=30$K. For source prompt tuning, the prompt token embeddings are initialized from sampled vocabulary (i.e., the 5,000 most common tokens). During target prompt tuning, 
we save a checkpoint every 500 steps and report results on the checkpoint with the highest validation performance. %
Appendix~\ref{appendix:training_details} contains training details for \ssc{PromptTuning} and model tuning approaches.
\subsection{Effect of \bspot} We compare the results of \spot and other approaches in Table~\ref{tbl:pretraining_results} and Figure~\ref{fig:superglue_graph}. Below, we summarize and analyze each of our findings in detail.
\paragraph{\bspot significantly improves performance and stability of \bssc{PromptTuning}:} Our results on the \ssc{GLUE} and \ssc{SuperGLUE} benchmarks with \ssc{T5 Base} (Table~\ref{tbl:pretraining_results}) suggest that prompt transfer provides an effective means of improving performance for \ssc{PromptTuning}. For example, the best-performing variant of \spot outperforms the vanilla \ssc{PromptTuning} approach on both \ssc{GLUE} and \ssc{SuperGLUE} by a substantial margin, obtaining +4.4 and +10.1 point %
average accuracy improvements, respectively. Our ablation study indicates that longer tuning is also an important ingredient for achieving our best performance, and is complementary to prompt transfer. Additionally, when longer tuning is omitted, we observe that \spot improves stability across runs.
\begin{table}[t]
\centering
\begin{adjustbox}{max width=0.48\textwidth}
\begin{tabular}{ l c c }
\toprule
\textbf{Method} & \bssc{GLUE} & \bssc{SuperGLUE} \\
\midrule
\midrule
\multicolumn{3}{l}{\ssc{Baseline}} \\ 
\hspace{10pt}\ssc{PromptTuning} & 81.2$_{\smallsup{0.4}}$ & 66.6$_{\smallsup{0.2}}$ \\ 
\hspace{20pt}$-$ longer tuning & 78.4$_{\smallsup{1.7}}$ & 63.1$_{\smallsup{1.1}}$ \\ 
\midrule
\midrule
\multicolumn{3}{l}{\spot \emph{with different source mixtures}} \\
\hspace{10pt}\ssc{GLUE} (\textit{8 tasks}) & \textbf{82.8}$_{\smallsup{0.2}}$ & \textbf{73.2}$_{\smallsup{0.3}}$ \\
\hspace{20pt}$-$ longer tuning & 82.0$_{\smallsup{0.2}}$ & 70.7$_{\smallsup{0.4}}$ \\
\midrule
\hspace{10pt}\ssc{C4}  & 82.0$_{\smallsup{0.2}}$ & 67.7$_{\smallsup{0.3}}$ \\
\hspace{10pt}\ssc{MNLI}  & 82.5$_{\smallsup{0.0}}$ & 72.6$_{\smallsup{0.8}}$ \\
\hspace{10pt}\ssc{SQuAD}  & 82.2$_{\smallsup{0.1}}$ & 72.0$_{\smallsup{0.4}}$ \\
\hspace{10pt}\ssc{SuperGLUE} (8 tasks) & 82.0$_{\smallsup{0.1}}$ & 66.6$_{\smallsup{0.2}}$ \\
\hspace{10pt}\ssc{NLI} (\textit{7 tasks}) & 82.6$_{\smallsup{0.1}}$ & 71.4$_{\smallsup{0.2}}$ \\
\hspace{10pt}Paraphrasing/similarity (\textit{4 tasks}) & 82.2$_{\smallsup{0.1}}$ & 69.7$_{\smallsup{0.5}}$ \\
\hspace{10pt}Sentiment (\textit{5 tasks}) & 81.1$_{\smallsup{0.2}}$ & 68.6$_{\smallsup{0.1}}$ \\
\hspace{10pt}\ssc{MRQA} (\textit{6 tasks}) & 81.8$_{\smallsup{0.2}}$ & 68.4$_{\smallsup{0.2}}$ \\
\hspace{10pt}\ssc{RAINBOW} (\textit{6 tasks}) & 80.3$_{\smallsup{0.6}}$ & 64.0$_{\smallsup{0.4}}$ \\
\hspace{10pt}Translation (\textit{3 tasks}) & 82.4$_{\smallsup{0.2}}$ & 65.3$_{\smallsup{0.1}}$ \\
\hspace{10pt}Summarization (\textit{9 tasks}) & 80.9$_{\smallsup{0.3}}$ & 67.1$_{\smallsup{1.0}}$ \\
\hspace{10pt}\ssc{GEM} (\textit{8 tasks}) & 81.9$_{\smallsup{0.2}}$ & 70.5$_{\smallsup{0.5}}$ \\
\hspace{10pt}All (C4 + 55 supervised tasks) & 81.8$_{\smallsup{0.2}}$ &  67.9$_{\smallsup{0.9}}$ \\
\midrule
\end{tabular}
\end{adjustbox}
\caption{\ssc{GLUE} and \ssc{SuperGLUE} results achieved by applying \ssc{T5 Base} with different prompt tuning approaches. We report the mean and standard deviation (in the subscript) across three random seeds. \spot significantly improves performance and stability of \ssc{PromptTuning} across the two benchmarks.}
\label{tbl:pretraining_results}
\vspace{-2mm}
\end{table}

Within \spot, we can compare the effectiveness of different source mixtures (see Table~\ref{tbl:pretraining_results}). Source prompt tuning on \ssc{GLUE} performs best on both \ssc{GLUE} and \ssc{SuperGLUE}, obtaining average scores of 82.8 and 73.2, respectively.\footnote{\fssc{SuperGLUE} tasks benefit less from source prompt tuning on \fssc{SuperGLUE} likely due to the small size of these datasets.} Interestingly, unsupervised source prompt tuning on \ssc{C4} %
(the same task used to pre-train our frozen models) still yields considerable improvements, even outperforming using \ssc{SuperGLUE} for \ssc{SuperGLUE} tasks. Using \ssc{MNLI} or \ssc{SQuAD} as a single source dataset is also particularly helpful across target tasks. Other source mixtures can lead to significant gains, with some families of tasks (e.g., \ssc{NLI} and paraphrasing/semantic similarity) showing more benefit than others. \aclcr{Mixing all the datasets together does not yield the best results, possibly due to task interference/negative transfer issues, where achieving good performance on one or more source tasks can hurt performance on a target task.}
\vspace{-1mm}
\paragraph{\bspot helps close the gap with \bssc{ModelTuning} across all model sizes:} 
Figure~\ref{fig:superglue_graph} shows our \ssc{SuperGLUE} results across model sizes (\aclcr{see Appendix~\ref{appendix:fig1_full_results} for full results}).
As shown in~\citet{BLester21}, \ssc{PromptTuning} becomes more competitive with scale, and at the \ssc{XXL} size, it nearly matches the performance of \ssc{ModelTuning}. %
However, at smaller model sizes, there are still large gaps between the two approaches. We show that \spot helps close these gaps and even exceeds \ssc{ModelTuning}'s performance by a large margin at several model sizes, while retaining all the computational benefits conferred by \ssc{PromptTuning}. Finally, at the \ssc{XXL} size, \spot achieves the best average score of 91.2, +1.1 points better than the strong \ssc{Multi-taskModelTuning} baseline, despite having 27,000$\times$ fewer task-specific parameters in both multi-task source tuning and target tuning.%

As a final test of \spot's effectiveness, we submitted our \ssc{XXL} model's predictions to the \ssc{SuperGLUE} leaderboard, achieving a score of \aclcr{89.2}. This far exceeds all previous submissions using parameter-efficient adaptation, such as \ssc{\mbox{GPT-3}}~(71.8), and almost matches fully fine-tuned \ssc{T5 XXL}~(89.3),\footnote{\aclcr{Note that the \ssc{T5} submission uses the original version of \ssc{T5} 
(which was pre-trained on a multi-task mixture of unsupervised and supervised tasks)
while we use \ssc{T5 1.1} (which was pre-trained on C4 only without mixing in supervised tasks).}} despite tuning 27,000$\times$ fewer parameters. \aclcr{To the best of our knowledge, \spot is the first parameter-efficient adaptation approach that is competitive with methods that tune billions of parameters.} See Appendix~\ref{appendix:superglue} for details. 
\section{Predicting task transferability}%
\label{sec:task-specific_spot}

So far, we have seen that soft prompt transfer can significantly boost the performance of prompt tuning, but it is critical to pick the right source tasks for transfer. For instance, through an extensive search, we found that \ssc{GLUE} and \ssc{MNLI} provide excellent source tasks for transferring to individual \ssc{GLUE} and \ssc{SuperGLUE} tasks. But what about a resource-constrained scenario where a user is not able to exhaustively search for a set of source tasks? Can we \emph{predict} which tasks will best transfer onto a novel target task without testing them one by one?

To investigate this, we conduct a large-scale empirical study with 26 \ssc{NLP} tasks. We first measure transferability across all task combinations (\S\ref{sec:measuring_transferability}). Next, we show that by interpreting task prompts as task embeddings, we can construct a semantic space of tasks, wherein similar tasks cluster together (\S\ref{sec:defining_similarity}). Based on this observation, we propose a retrieval algorithm (\S\ref{sec:predict_and_exploit}) that leverages task embedding similarity to choose which source tasks to use for a given novel target task (Figure~\ref{fig:approach}, right). Our proposed approach can eliminate $69$\% of the source task search space while keeping $90$\% of the best-case quality gain. 

\subsection{Measuring transferability}
\label{sec:measuring_transferability}

\begin{table}[t!]
\centering
\footnotesize
\begin{adjustbox}{max width=0.48\textwidth}
\begin{tabular}{l l r}
\toprule
\multicolumn{1}{l}{\textbf{Name}} & \multicolumn{1}{l}{\textbf{Task type}} & \multicolumn{1}{r}{\textbf{$|$Train$|$}} \\ [0.5ex]
\multicolumn{3}{l}{\emph{16 source tasks}} \\ [0.5ex]
\ssc{C4} & language modeling & 365M\\
\ssc{DocNLI} & \ssc{NLI} & 942K \\
\ssc{Yelp-2} & sentiment analysis & 560K \\
\ssc{MNLI} & \ssc{NLI} & 393K \\
\ssc{QQP} &  paraphrase detection & 364K \\
\ssc{QNLI} & \ssc{NLI} & 105K \\
\ssc{ReCoRD} & \ssc{QA} & 101K \\
\ssc{CxC} & semantic similarity & 88K \\
\ssc{SQuAD} & \ssc{QA} & 88K \\
\ssc{DROP} & \ssc{QA} & 77K \\
\ssc{SST-2} & sentiment analysis & 67K \\
\ssc{WinoGrande} & commonsense reasoning & 40K \\
\ssc{HellaSWAG} & commonsense reasoning & 40K \\
\ssc{MultiRC} & \ssc{QA} & 27K \\
\ssc{CosmosQA} & commonsense reasoning & 25K \\
\ssc{RACE} & \ssc{QA} & 25K \\
\midrule 
\multicolumn{3}{l}{\emph{10 target tasks}} \\ [0.5ex] 
\ssc{BoolQ} & \ssc{QA} & 9K \\
\ssc{CoLA} & grammatical acceptability & 9K \\
\ssc{STS-B} & semantic similarity & 6K \\
\ssc{WiC} & word sense disambiguation & 5K \\
\ssc{CR} & sentiment analysis & 4K \\
\ssc{MRPC} & paraphrase detection & 4K \\
\ssc{RTE} & \ssc{NLI} & 2K \\
\ssc{WSC} & coreference resolution & 554 \\
\ssc{COPA} & \ssc{QA} & 400 \\
\ssc{CB} & \ssc{NLI} & 250 \\
\bottomrule
\end{tabular}
\end{adjustbox}
\caption{Tasks used in our task transferability experiments, sorted by training dataset size.} 
\label{tbl:transfer_datasets}
\vspace*{-2mm}
\end{table}

We study a diverse set of 16 source datasets and 10 target datasets (see Table~\ref{tbl:transfer_datasets}).\footnote{Beyond the datasets from \S\ref{sec:prompt_pretraining}, we use \fssc{DocNLI}~\cite{WYin21}, \fssc{Yelp-2}~\cite{XZhang15}, \fssc{CxC}~\cite{ZParekh21}, \fssc{DROP}~\cite{DDua19}, \fssc{WinoGrande}~\cite{KSakaguchi20}, \fssc{HellaSWAG}~\cite{RZellers19}, \fssc{CosmosQA}~\cite{LHuang19}, \fssc{RACE}~\cite{GLai17}, and \fssc{CR}~\cite{MHu04}.}  
We consider all 160 possible source-target pairs, and perform transfer from each source task to each target task. All source tasks are data-rich or have been shown to yield positive transfer in prior work. To simulate a realistic scenario, we use low-resource tasks (less than 10K training examples) as target tasks.\footnote{The source tasks comprise one unsupervised task (\fssc{C4}) and 15 supervised tasks covering natural language inference (\fssc{NLI}), paraphrasing/semantic similarity, sentiment analysis, question answering (\fssc{QA}), and commonsense reasoning. The target tasks additionally include grammatical acceptability, word sense disambiguation, and coreference resolution.} 

To limit computational costs, we use \ssc{T5 Base} in all of our task transferability experiments. We perform $262{,}144$ prompt tuning steps on each source task. The prompt checkpoint with the highest source task validation performance is selected to initialize prompts for different target tasks. Since the target datasets are small, we only perform 100K prompt tuning steps on each target task. We repeat each experiment three times with different random seeds. Other training details match~\S\ref{sec:training_details}.
\begin{figure}[t!]
\centering
\includegraphics[width=0.48\textwidth]{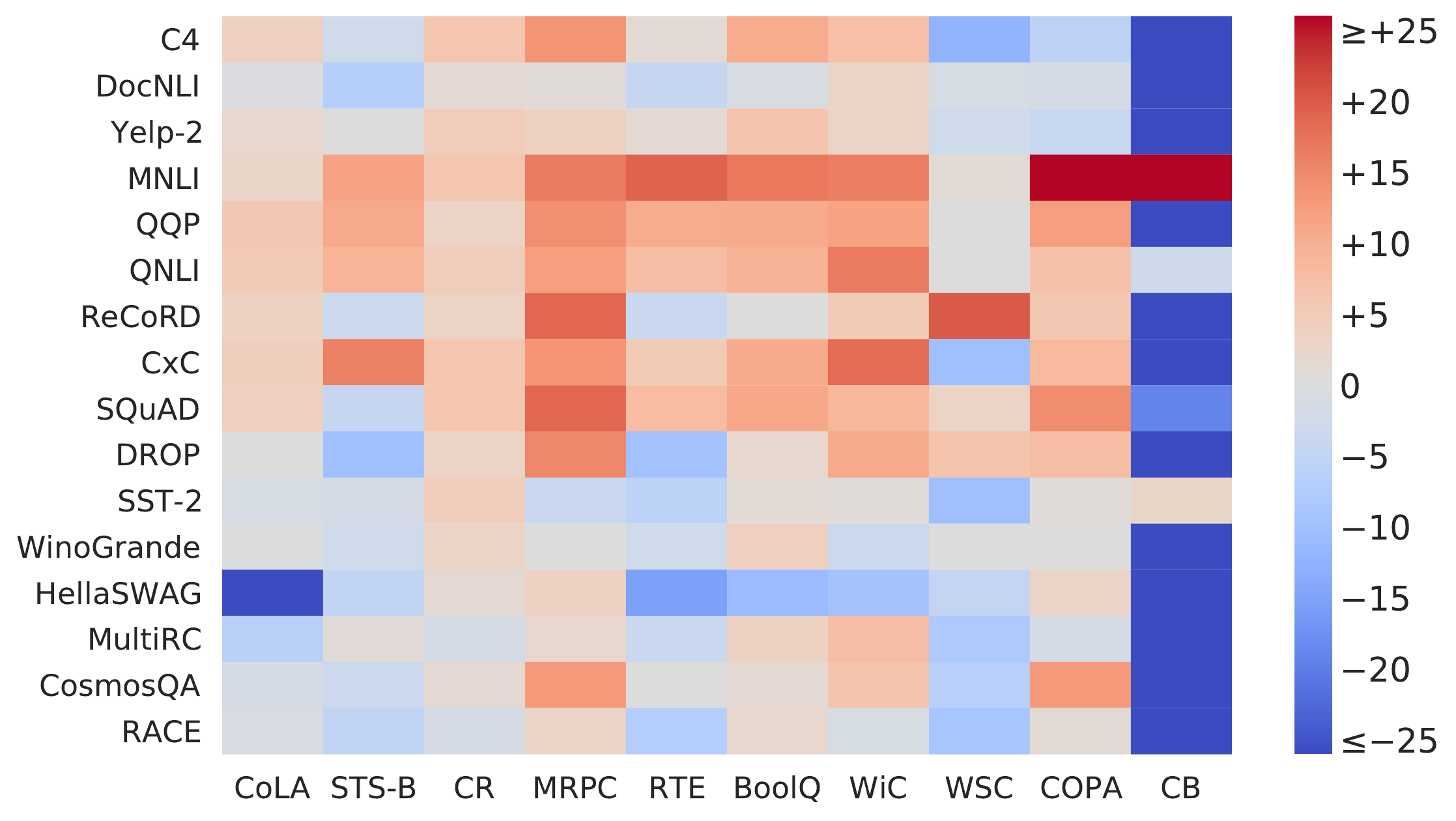}
\caption{%
A heatmap of our task transferability results. Each cell shows the relative error reduction on the target task of the transferred prompt from the associated source task (row) to the associated target task (column).}
\label{fig:transfer_heat_map}
\end{figure}
\paragraph{Tasks can benefit each other via prompt transfer:} Figure~\ref{fig:transfer_heat_map} shows a heatmap of our results (see Appendix~\ref{appendix:transfer_results} for full results). In many cases, prompt transfer provides a significant gain on the target task. The transfer \ssc{MNLI}$\rightarrow$ \ssc{CB} yields the largest relative error reduction of 58.9\% (from an average score of 92.7 to 97.0), followed by \ssc{MNLI}$\rightarrow$ \ssc{COPA} (29.1\%) and \ssc{ReCoRD}$\rightarrow$ \ssc{WSC} (20.0\%). Using the best source prompt (out of 48) for each target task dramatically improves the average score across 10 target tasks from 74.7 to 80.7. Overall, our results show effective transfer from large source tasks that involve high-level reasoning about semantic relationships among sentences (e.g., \ssc{MNLI}), or when the source and target tasks are similar (e.g., \ssc{CxC}$\rightarrow$ \ssc{STS-B}). Interestingly, positive transfer can occur between relatively dissimilar tasks (e.g., \ssc{ReCoRD}$\rightarrow$ \ssc{WSC}, \ssc{SQuAD}$\rightarrow$ \ssc{MRPC}, \ssc{CxC}$\rightarrow$ \ssc{WiC}).\footnote{Table~\ref{effective_transfers} in Appendix~\ref{appendix:transfer_results} contains more cases.} %
\subsection{Defining task similarity through prompts}
\label{sec:defining_similarity}
\begin{figure}[t!]
\centering
\includegraphics[width=0.48\textwidth]{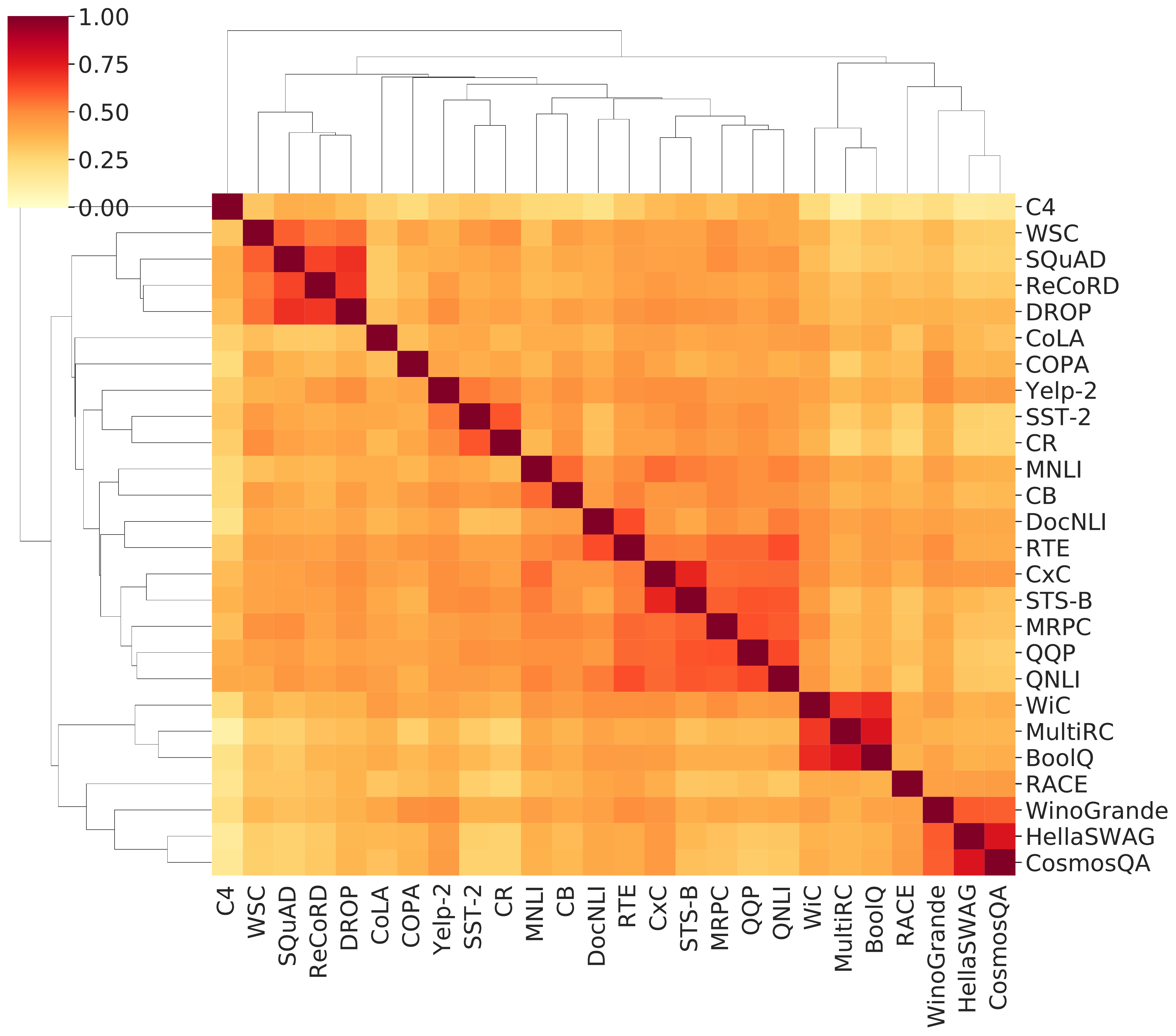}
\caption{%
A clustered heatmap of cosine similarities between the task embeddings of the 26 NLP tasks we study. Our prompt-based task embeddings capture task relationships: similar tasks cluster together.}
\vspace{-2mm}
\label{fig:task_clustermap}
\end{figure}
Since only prompt parameters are updated during prompt tuning on specific tasks, the learned prompts likely encode task-specific knowledge. This suggests that they could be used to reason about the nature of tasks and their relationships. To test this idea, we interpret task prompts as \textit{task embeddings} and construct a semantic space of tasks. %
More concretely, we define a task's embedding as the prompt checkpoint after training for 10K steps on that task.\footnote{Our preliminary experiments with other checkpoint alternatives (in the range 1K to 100K) yielded worse performance. \aclcr{We also found that measuring task similarity using task embeddings derived from a \emph{fixed} prompt checkpoint (10K steps) gave better results than those derived from the \emph{best-performing} prompt checkpoint per task. This suggests that prompts trained for a differing number of steps may be less directly comparable than those trained for the same length.}} %
Note that using early checkpoints allows for quick computation of task embeddings for novel target tasks. %
We estimate the similarity between two tasks $t^{1}, t^{2}$ by measuring the similarity between their corresponding task embeddings $\bm{e}^{1}, \bm{e}^{2}$, using the following metrics:
\paragraph{\bssc{Cosine Similarity of Average Tokens}:} We compute the cosine similarity between the average pooled representations of the prompt tokens: 
\[sim(t^1, t^2) = cos(\dfrac{1}{\bmmc{L}}\sum_{i} \bm{e}_i^{1}, \dfrac{1}{\bmmc{L}}\sum_{j} \bm{e}_j^{2}),\] where $\bm{e}_i^{1}, \bm{e}_j^{2}$ denote the respective prompt tokens of $\bm{e}^{1}, \bm{e}^{2}$, and $cos$ denotes the cosine similarity.
\paragraph{\bssc{Per-token Average Cosine Similarity}:} We compute the average cosine similarity between every prompt token pair $(\bm{e}_i^{1}, \bm{e}_j^{2})$:
\[sim(t^1, t^2) = \dfrac{1}{\bmmc{L}^2} \sum_{i} \sum_{j}  cos(\bm{e}_i^{1}, \bm{e}_j^{2}).\]
\paragraph{Task embeddings capture task relationships:} Figure~\ref{fig:task_clustermap} shows a hierarchically-clustered heatmap of cosine similarities between the task embeddings using the \ssc{Cosine Similarity of Average Tokens} metric.\footnote{To obtain the highest resolution of similarity between two tasks, we use the average of cosine similarities between their task embeddings obtained with all the three different prompt tuning runs (9 combinations).} We observe that our learned task embeddings capture many intuitive task relationships. Specifically, similar tasks group together into clusters, including \ssc{QA} (\ssc{SQuAD}, \ssc{ReCoRD}, and \ssc{DROP}; \ssc{MultiRC} and \ssc{BoolQ}), sentiment analysis (\ssc{\mbox{Yelp-2}}, \ssc{\mbox{SST-2}}, and \ssc{CR}), \ssc{NLI} (\ssc{MNLI} and \ssc{CB}; \ssc{DocNLI} and \ssc{RTE}), semantic similarity (\ssc{\mbox{STS-B}} and \ssc{CxC}), paraphrasing (\ssc{MRPC} and \ssc{QQP}), and commonsense reasoning (\ssc{WinoGrande}, \ssc{HellaSWAG}, and \ssc{CosmosQA}). We note that \ssc{QNLI}, which is an \ssc{NLI} task built from the \ssc{SQuAD} dataset, is not closely linked to \ssc{SQuAD}; this suggests that our task embeddings are more sensitive to the type of task than domain similarity. %
Interestingly, they also capture the unintuitive case of \ssc{ReCoRD}'s high transferability to \ssc{WSC}. Additionally, task embeddings that are derived from different prompts of the same task have high similarity scores (see Appendix~\ref{appendix:task_embedding_similarity}).

\subsection{Predicting transferability via similarity}
\label{sec:predict_and_exploit}
We leverage our task embeddings to predict and exploit task transferability. Specifically, we explore methods to predict the most beneficial source tasks for a given target task and then make use of their prompts to improve performance on the target task. To enlarge our set of source prompts, we use the prompts from each of the three different prompt tuning runs on each source task, resulting in 48 source prompts. Given a target task $t$ with task embedding $\bm{e}^t$, we rank all the source prompts $\bm{\rho}^{s}$ with associated embeddings $\bm{e}^{s}$ in descending order by the similarity $sim(\bm{e}^{s}, \bm{e}^{t})$. We denote the ranked list of source prompts as $\bm{\rho}^{s_r}$, where $r$ denotes the rank $(r=1,2,\ldots, 48)$. We experiment with three methods for exploiting the ranked source prompts:%

\paragraph{\bssc{Best of Top-$k$}:} We select the top-$k$ source prompts and use each of them individually to initialize the target prompt. This procedure requires prompt tuning $k$ times on the target task $t$.
The best individual result is used for evaluating the effectiveness of this method.

\paragraph{\bssc{Top-$k$ Weighted Average}:} We initialize the target prompt with a weighted average of the \mbox{top-$k$} source prompts $\sum_{r=1}^{k} \alpha_{r} \bm{\rho}^{s_r}$ so that we only perform prompt tuning on the target task $t$ once. The weights $\alpha_{r}$ are computed as:
\[\alpha_{r} = \dfrac{sim(\bm{e}^{s_r}, \bm{e}^{t})}{\sum_{l=1}^{k} sim(\bm{e}^{s_l}, \bm{e}^{t})},
\]
where $\bm{e}^{s_r}$ denotes the corresponding task embedding of $\bm{\rho}^{s_r}$.
\begin{figure}[t!]
\centering
\includegraphics[width=0.48\textwidth]{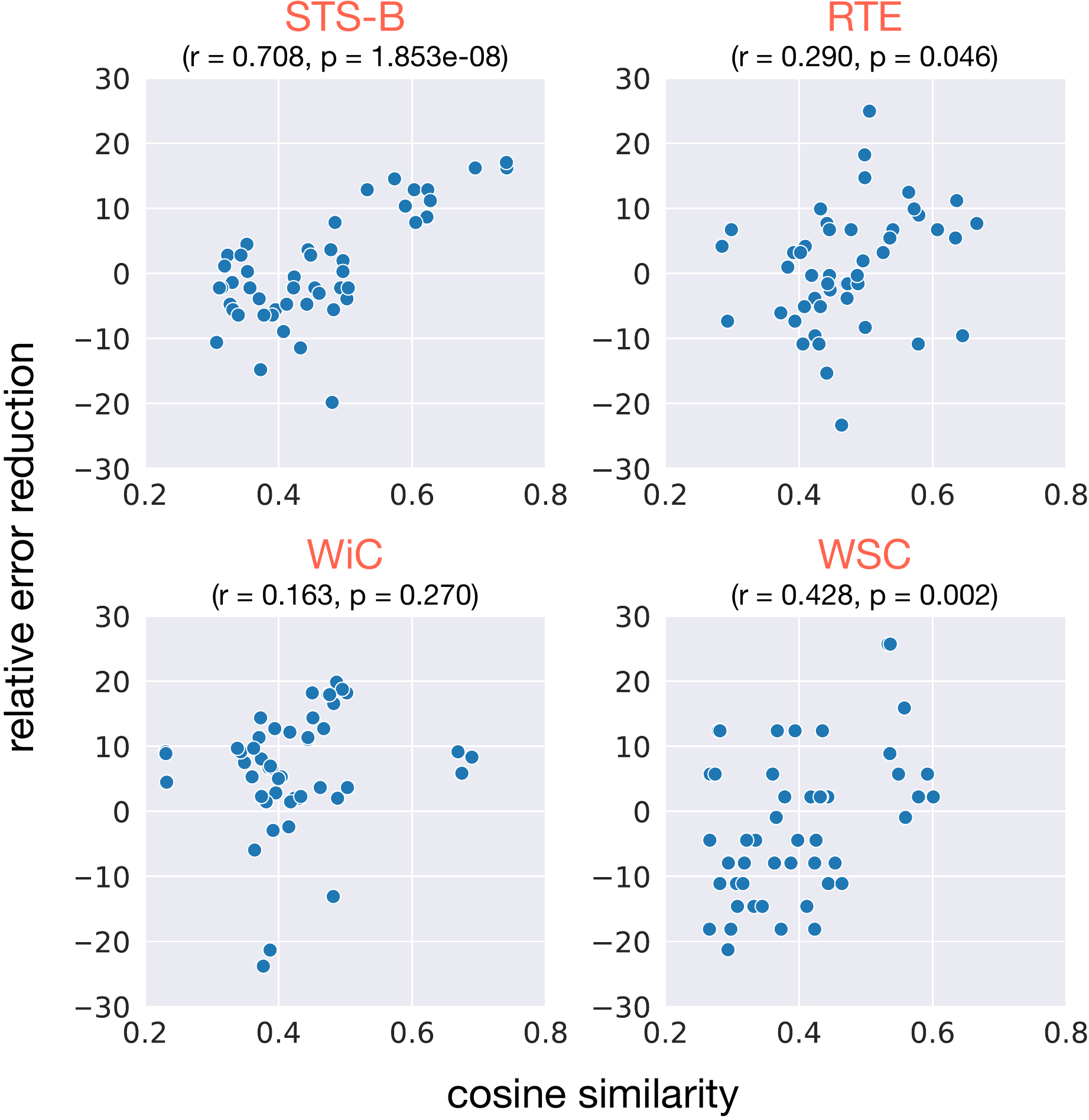}
\caption{Correlation between task similarity and task transferability. Each point represents a source prompt. The x-axis shows the cosine similarity between the associated source and target task embeddings, averaged over three runs for the target task (orange title). The y-axis measures the relative error reduction on the target task achieved by each source prompt. We include the Pearson correlation coefficient ($r$) and p-value.} %
\label{fig:similarity_vs_transferability}
\vspace{-3mm}
\end{figure}

\paragraph{\bssc{Top-$k$ Multi-task Mixture}:} We first identify the source tasks whose prompts are in the \mbox{top-$k$} prompts and mix their datasets and the target dataset together, using the examples-proportional mixing strategy of~\citet{CRaffel20}. Then, we perform source prompt tuning on this multi-task mixture and use the final prompt checkpoint to initialize the prompt for target prompt tuning.

We report the average score across all target tasks achieved by each method. For comparison, we measure the absolute and relative improvements over \ssc{Baseline}---prompt tuning on each target task from scratch (i.e., without any prompt transfer).\footnote{For each target task $t$, we report the average and standard deviation of performance across three prompt tuning runs.} Additionally, we include \ssc{Oracle}---the oracle results achieved by a brute-force search to identify the best possible out of 48 source prompts for each target task.
\paragraph{Correlation between task similarity
and task transferability:}
Figure~\ref{fig:similarity_vs_transferability} shows how the relative error reduction on a target task changes as a function of the similarity between the source and target task embeddings. Overall, we observe a significant positive correlation between task embedding similarity and task transferability on four (out of 10) target tasks, including \ssc{STS-B} ($p < 0.001$), \ssc{CB} ($p < 0.001$), \ssc{WSC} ($p < 0.01$), and \ssc{RTE} ($p < 0.05$), while it is less significant on the other tasks.\footnote{See Appendix~\ref{appendix:correlation_similarity_transferability} for full results.} \aclcr{In some cases (e.g., on \ssc{BoolQ}), we observe a large relative error reduction (19.0\%, achieved by a source prompt of \ssc{MNLI}) despite a low cosine similarity (0.4). This suggests that factors other than task similarity (data size, task difficulty, domain similarity, etc.)\@ may also play a role in determining transferability.}
\paragraph{Retrieving targeted source tasks via task embeddings is helpful:}
Table~\ref{tbl:predict_transferability} compares different methods for identifying which source prompts could be beneficial for a given target task. Overall, 
our results show the effectiveness of \ssc{Best of Top-$k$}: 
simply choosing the source prompt with the highest task embedding similarity to the target task using  \ssc{Per-token Average Cosine Similarity} improves over the baseline by a large margin (from an average score of 74.7 to 76.7, a 12.1\% average relative error reduction).
Trying all the top-3 (out of 48) source prompts for each target task yields an average score of 77.5.
With larger values of $k$, we can retain most of the benefits of oracle selection %
(80\% of the gain in terms of average score with $k=9$ and 90\% with $k=15$), while still eliminating over 2/3 of the candidate source prompts. \ssc{Top-$k$ Weighted Average} has similar average performance to \ssc{Best of Top-$k$} with $k=1$, but achieves lower variance. Thus, this may be an appealing alternative to \ssc{Best of Top-$k$} in scenarios where trying multiple prompt tuning runs on the target task is prohibited. Finally, \ssc{Top-$k$ Multi-task Mixture} also provides a means of obtaining strong performance with an average score of 77.8, even outperforming \ssc{Best of Top-$k$} with $k \leq 3$.

\begin{table}[t!]
\centering
\footnotesize
\begin{adjustbox}{max width=0.48\textwidth}
\begin{tabular}{ l c c c}
\toprule
\multirow{2}{*}{\textbf{Method}} & \multicolumn{2}{c}{\textbf{Change}} & \multirow{2}{*}{\textbf{Avg. score}} \\
\cmidrule(l){2-3} 
& \textbf{Abs.} & \textbf{Rel.} & \\
\midrule
\midrule
\ssc{Baseline} & - & - & 74.7$_{\smallsup{0.7}}$ \\
\midrule
\midrule
\multicolumn{4}{l}{\ssc{Brute-force Search} ($k=48$)} \\
\ssc{Oracle} & 6.0$_{\smallsup{0.5}}$ &
26.5$_{\smallsup{1.1}}$ & 80.7$_{\smallsup{0.0}}$ \\
\midrule
\midrule
\multicolumn{4}{l}{\ssc{Cosine Similarity of Average Tokens}} \\
\textbf{\ssc{Best of Top-$k$}} \\
\hspace{10pt}$k=1$ & 1.5$_{\smallsup{0.5}}$ & 11.7$_{\smallsup{1.1}}$ & 76.2$_{\smallsup{0.1}}$ \\
\hspace{10pt}$k=3$ & 2.7$_{\smallsup{0.6}}$ & 16.6$_{\smallsup{1.1}}$ & 77.4$_{\smallsup{0.3}}$ \\
\hspace{10pt}$k=6$ & 3.8$_{\smallsup{0.1}}$ & 20.0$_{\smallsup{1.1}}$ & 78.5$_{\smallsup{0.5}}$ \\
\hspace{10pt}$k=9$ & 4.5$_{\smallsup{0.4}}$ & 22.2$_{\smallsup{1.1}}$ & 79.2 $_{\smallsup{0.1}}$ \\
\hspace{10pt}$k=12$ & 5.0$_{\smallsup{0.9}}$ & 23.6$_{\smallsup{2.2}}$ & 79.7 $_{\smallsup{0.4}}$ \\
\hspace{10pt}$k=15$ & 5.4$_{\smallsup{0.8}}$ & 24.9$_{\smallsup{1.8}}$ & 80.1$_{\smallsup{0.3}}$ \\
\midrule
\midrule
\multicolumn{4}{l}{\ssc{Per-token Average Cosine Similarity}} \\
\textbf{\ssc{Best of Top-$k$}} \\
\hspace{10pt}$k=1$ & 2.0$_{\smallsup{0.4}}$ & 12.1$_{\smallsup{1.1}}$ & 76.7$_{\smallsup{0.7}}$ \\
\hspace{10pt}$k=3$ & 2.9$_{\smallsup{0.6}}$ & 17.0$_{\smallsup{0.6}}$ & 77.5$_{\smallsup{0.4}}$ \\
\hspace{10pt}$k=6$ & 4.5$_{\smallsup{0.5}}$ & 22.1$_{\smallsup{1.2}}$ & 79.2$_{\smallsup{0.1}}$ \\
\hspace{10pt}$k=9$ & 4.6$_{\smallsup{0.5}}$ & 22.6$_{\smallsup{0.9}}$ & 79.5$_{\smallsup{0.2}}$ \\
\hspace{10pt}$k=12$ & 5.0$_{\smallsup{0.6}}$ & 23.5$_{\smallsup{1.4}}$ & 79.6$_{\smallsup{0.1}}$ \\
\hspace{10pt}$k=15$ & 5.3$_{\smallsup{0.9}}$ & 24.5$_{\smallsup{2.2}}$ & 80.0$_{\smallsup{0.4}}$ \\
\midrule
\ssc{Top-$k$ Weighted Average} \\
\hspace{10pt}best $k=3$ & 1.9$_{\smallsup{0.5}}$ & 11.5$_{\smallsup{2.7}}$ & 76.6$_{\smallsup{0.1}}$ \\
\midrule
\ssc{Top-$k$ Multi-task Mixture} \\
\hspace{10pt}best $k=12$ & 3.1$_{\smallsup{0.5}}$ & 15.3$_{\smallsup{2.8}}$ & 77.8$_{\smallsup{0.1}}$ \\
\bottomrule
\end{tabular}
\end{adjustbox}
\caption{Task embeddings provide an effective means of predicting and exploiting task transferability. Using \ssc{Best of Top-$k$} with $k=3$ improves over \ssc{Baseline} (\ssc{PromptTuning} on
each task from scratch) by +2.8 points. With larger values of $k$ ($\leq 15$), we can retain most of the benefits conferred by oracle selection. For \ssc{Top-$k$ Weighted Average} and \ssc{Top-$k$ Multi-task Mixture}, we experiment with different values of $k \in \{3,6,9,12\}$ and report the best results.}%
\label{tbl:predict_transferability}
\vspace{-3mm}
\end{table}

\section{Related Work}
\label{section:related_work}
\paragraph{Parameter-efficient transfer learning:} Large-scale pre-trained language models have been shown to exhibit remarkable performance on many NLP tasks~\cite{JDevlin19,YLiu19,ZYang19,ZLan20,CRaffel20,TBrown20,PHe21}.
%
To improve practical applicability of these models, early work uses compression techniques~\cite{VSanh19,XJiao20,AFan20,VSanh20}
to obtain lightweight models. Other work involves updating only small parts of the model~\cite{EZaken21} or task-specific modules, such as adapters~\cite{NHoulsby19,RMahabadi21a} or low-rank structures~\cite{RMahabadi21b,EHu21},
while keeping the rest of the model fixed.

Recently, \citet{TBrown20} demonstrate impressive few-shot performance 
with \ssc{PromptDesign}, where their model is conditioned on a manual text prompt at inference time to perform different tasks. Several efforts have since focused on developing prompt-based learning approaches with carefully handcrafted prompts~\cite{TSchick20}%
, prompt mining and paraphrasing~\cite{ZJiang20}, gradient-based search for improved prompts~\cite{TShin20}, and automatic prompt generation%
~\cite{TGao21}. The use of hard prompts, however, was found to be sub-optimal and sensitive to the choice of the prompt~\cite{TZhao21,XLiu21}. 
As such, more recent work has shifted toward learning soft prompts~\cite{XLiu21,GQin21,XLi21,BLester21}, which can be seen as learnable parameters injected into the model. We refer readers to \citet{PLiu21} for a recent survey on prompt-based learning research. 

In concurrent work, \citet{YGu21} also explore the effectiveness of prompt transfer. Their method uses hand-crafted pre-training tasks tailored to specific types of downstream task, and thus may be less extensible to novel downstream tasks. In contrast, we use existing tasks as source tasks and show that prompt transfer can confer benefits even when there are mismatches (e.g., in task type or input/output format) between the source and target. 
\paragraph{Task transferability}
We also build on existing work on task transferability~\cite{AWang19c,NLiu19,ATalmor19,YPruksachatkun20,TVu20,TVu21}. Prior work shows effective transfer from data-rich source tasks~\cite{JPhang19}, those that require complex reasoning and inference~\cite{YPruksachatkun20}, or those that are similar to the target task~\cite{TVu20}. There have also been efforts to predict task transferability~\cite{JBingel17,TVu20,CPoth21}. \citet{TVu20} use task embeddings derived from either the input text or the diagonal Fisher information matrix of the model, while \citet{CPoth21} explore adapter-based alternatives. Here, our use of the same model (without task-specific components) and a unified text-to-text format allows us to better model the space of tasks. Additionally, prompt-based task embeddings are comparatively cheaper to obtain. %

\section{Limitations \& Future work}
\aclcr{
As other parameter-efficient adaptation methods (see \S\ref{section:related_work}) may outperform \ssc{PromptTuning} in specific situations, it would be interesting to test whether an approach similar to \spot could extend successfully to these methods. At the same time, we believe that \ssc{PromptTuning} has its own merit. As pre-trained language models become larger and larger, some advantages of \ssc{PromptTuning} over other methods are: (1) Among current methods with learnable parameters, \ssc{PromptTuning} is the most parameter efficient, requiring less than 0.01\% task-specific parameters for most model sizes. (2) \ssc{PromptTuning} is simpler than other methods, as it does not modify the internal model architecture (cf.~the \ssc{Prefix-Tuning} method of \citet{XLi21}, which adds a prefix to each layer of both the Transformer encoder and decoder); as such, \ssc{PromptTuning} allows mixed-task inference and facilitates transfer learning between tasks. (3) As model capacity increases, \ssc{PromptTuning} becomes more competitive with \ssc{ModelTuning}; to the best of our knowledge, this has not been shown for other methods. (4) Soft prompts could possibly be interpreted as natural language instructions.
}


\aclcr{Additionally, since our prompt-based task embedding approach does not capture all of the factors that influence task transferability, we leave further exploration of other task embedding methods to future work.
}
\section{Conclusion}
\label{section:conclusion}

In this paper, we study transfer learning in the context of prompt tuning. We show that scale is not necessary for \ssc{PromptTuning} to match the performance of \ssc{ModelTuning}. \aclcr{On \ssc{SuperGLUE},} our \spot approach matches or even exceeds the performance of \ssc{ModelTuning} by a large margin across model sizes %
while being more parameter-efficient. %
Our large-scale study on task transferability indicates that tasks can benefit each other via prompt transfer in various scenarios. Finally, we demonstrate that task prompts can be interpreted as task embeddings to formalize the similarity between tasks. We propose a simple yet efficient retrieval approach that measures task  similarity to identify which source tasks could confer benefits to a novel target task. Taken as a whole, we hope that our work will spur more research into prompt-based transfer learning.
\section*{Acknowledgements}
We thank Mohit Iyyer, Sebastian Ruder, Kalpesh Krishna, Thang Luong, Quoc Le, and the members of the Descartes team and the UMass NLP group for helpful discussion and feedback. We would also like to thank Grady Simon, Lucas Dixon, Slav Petrov, Nader Akoury, Haw-Shiuan Chang, Katherine Thai, Marzena Karpinska, and Shufan Wang for their comments on this manuscript. Finally, we are grateful to Vamsi Aribandi for his work on preprocessing several datasets used in our experiments.

\bibliography{custom}
\bibliographystyle{acl_natbib}

\appendix
\clearpage
\newpage
\appendix
\section*{Appendices}
\label{section:appendices}
\section{Full results for Figure~\ref{fig:superglue_graph}}
Table~\ref{tbl:fig1} shows the performance of different model tuning and prompt tuning methods (described in~\S\ref{baselines}) on the \ssc{SuperGLUE} benchmark.
\label{appendix:fig1_full_results}

\section{Source datasets used in our \bspot experiments in~\S\ref{sec:prompt_pretraining}}
\label{appendix:source_datasets}
\begin{figure*}[t!]
\centering
\includegraphics[width=0.80\textwidth]{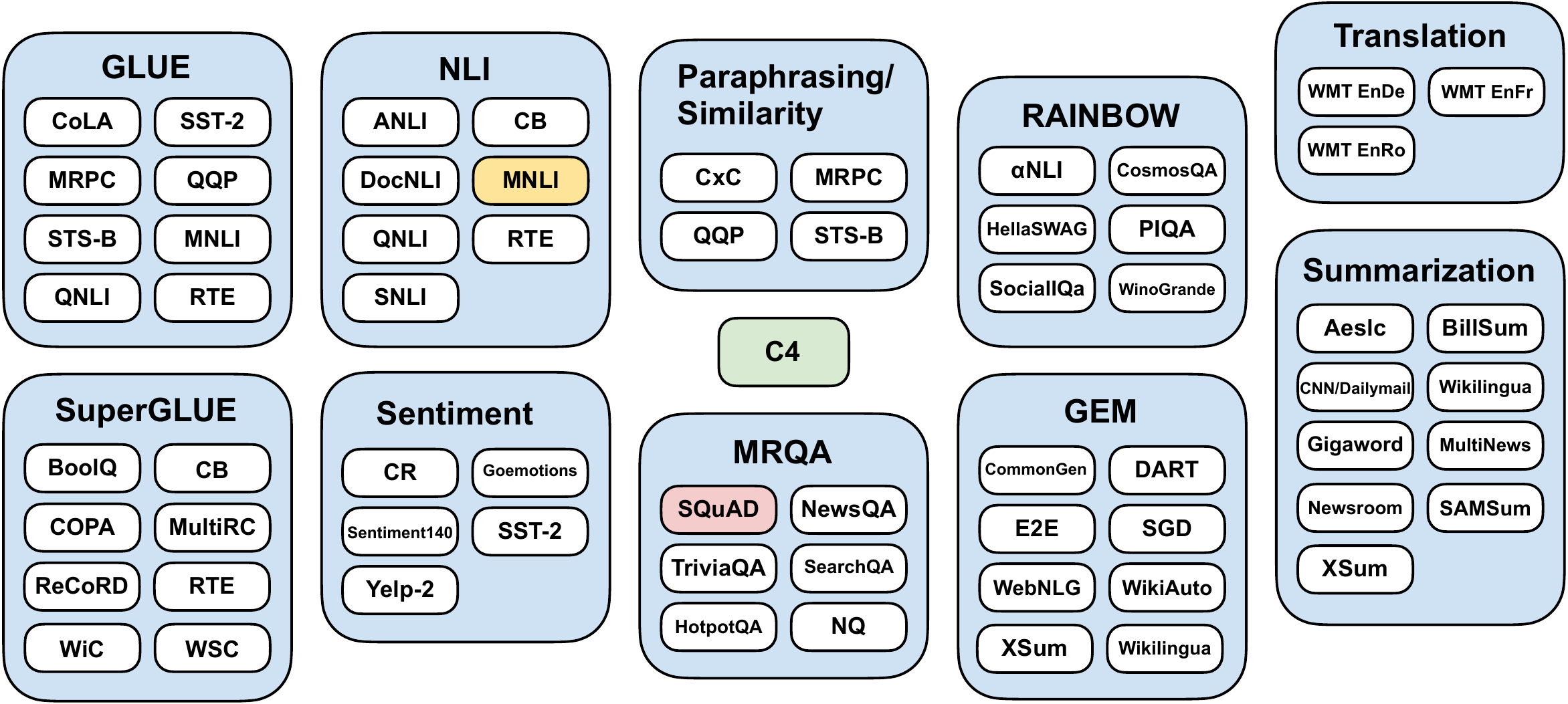}
\caption{Datasets used in our \spot experiments in~\S\ref{sec:prompt_pretraining}. \ssc{C4}, \ssc{MNLI}, and \ssc{SQuAD} were all used by themselves as single source tasks in addition to being mixed in with other tasks.}
\label{fig:prompt_pretraining_data}
\vspace{-2mm}
\end{figure*}
Figure~\ref{fig:prompt_pretraining_data} displays the datasets used in our \spot experiments in~\S\ref{sec:prompt_pretraining}. In addition to the \ssc{C4} unlabeled dataset~\cite{CRaffel20}, we use 55 labeled datasets. These datasets come from common \ssc{NLP} benchmarks/families of tasks, namely: 
\begin{itemize}
    \item \ssc{GLUE}~\cite{AWang19a}, including \ssc{CoLA}~\cite{AWarstadt19}, \ssc{SST-2}~\cite{RSocher13}, \ssc{MRPC}~\cite{WDolan05}, \ssc{QQP}~\cite{SIyer17}, \ssc{STS-B}~\cite{DCer17}, \ssc{MNLI}~\cite{AWilliams18}, \ssc{QNLI}~\cite{AWang19a}, and \ssc{RTE}~\cite[et seq.]{IDagan06}.
    
    \item \ssc{SuperGLUE}~\cite{AWang19b}, including \ssc{BoolQ}~\cite{CClark19}, \ssc{CB}~\cite{MDeMarneffe19}, \ssc{COPA}~\cite{MRoemmele11}, \ssc{MultiRC}~\cite{DKhashabi18}, \ssc{ReCoRD}~\cite{SZhang18}, \ssc{RTE}, \ssc{WiC}~\cite{MPilehvar19}, and \ssc{WSC}~\cite{HLevesque12}.
    
    \item Natural language inference (\ssc{NLI}), including \ssc{ANLI}~\cite{YNie20}, \ssc{CB}, \ssc{DocNLI}~\cite{WYin21}, \ssc{MNLI}, \ssc{QNLI}, \ssc{RTE}, and \ssc{SNLI}~\cite{SBowman15}.
    
    \item Paraphrasing/semantic similarity, including \ssc{CxC}~\cite{ZParekh21}, \ssc{MRPC}, \ssc{QQP}, and \ssc{STS-B}.
    
    \item Sentiment analysis, including \ssc{CR}~\cite{MHu04}, \ssc{Goemotions}~\cite{DDemszky20}, \ssc{Sentiment140}~\cite{AGo09}, \ssc{SST-2}, and \ssc{Yelp-2}~\cite{XZhang15}. 
    
    \item Question answering (\ssc{QA}) on \ssc{MRQA}~\cite{AFisch19}, including \ssc{SQuAD}~\cite{PRajpurkar16}, \ssc{NewsQA}~\cite{ATrischler17}, \ssc{TriviaQA}~\cite{MJoshi17}, \ssc{SearchQA}~\cite{MDunn17}, \ssc{HotpotQA}~\cite{ZYang18}, and \ssc{NaturalQuestions} (\ssc{NQ}~\cite{TKwiatkowski19}).
    
    \item Commonsense reasoning on \ssc{RAINBOW}~\cite{NLourie21} including $\alpha$\ssc{NLI}~\cite{CBhagavatula20}, \ssc{CosmosQA}~\cite{LHuang19}, \ssc{HellaSWAG}~\cite{RZellers19}, \ssc{PIQA}~\cite{YBisk20}, \ssc{SocialIQa}~\cite{MSap19}, and \ssc{WinoGrande}~\cite{KSakaguchi20}.
    
    \item Machine translation, including \ssc{WMT EnDe}~\cite{OBojar14}, \ssc{WMT EnFr}~\cite{OBojar15}, and \ssc{WMT EnRo}~\cite{OBojar16}.
    
    \item Summarization, including \ssc{Aeslc}~\cite{RZhang19}, \ssc{BillSum}~\cite{AKornilova19}, \ssc{CNN/Dailymail}~\cite{KHermann15,ASee17}, \ssc{Wikilingua}~\cite{FLadhak20}, \ssc{Gigaword}~\cite{DGraff03,ARush15}, \ssc{MultiNews}~\cite{AFabbri19}, \ssc{Newsroom}~\cite{MGrusky18}, \ssc{SAMSum}~\cite{BGliwa19}, and \ssc{XSum}~\cite{SNarayan18}.
    
    \item Natural language generation on \ssc{GEM}~\cite{SGehrmann21}, including \ssc{CommonGen}~\cite{YLin20}, \ssc{DART}~\cite{LNan21}, \ssc{E2E}~\cite{ODusek19}, \ssc{SGD}~\cite{ARastogi20}, \ssc{WebNLG}~\cite{CGardent17}, \ssc{WikiAuto}~\cite{CJiang20}, \ssc{XSum}, and \ssc{Wikilingua}.
\end{itemize}
\begin{table}[t!]
\centering
\footnotesize
\begin{adjustbox}{max width=0.48\textwidth}
\begin{tabular}{ l c c c c c}
\toprule
\multirow{2}{*}{\textbf{Method}} & \multicolumn{5}{c}{\textbf{Model size}} \\
\cmidrule(l){2-6} 
& \ssc{Small} & \ssc{Base} & \ssc{Large} & \ssc{XL} & \ssc{XXL} \\
\midrule
\midrule
\ssc{PromptDesign} (\ssc{\mbox{GPT-3}})  & 40.6 & 43.4 & 45.1 & 47.8 & 52.8 \\
\ssc{ModelTuning} & 62.8$_{\smallsup{0.8}}$ & 73.7$_{\smallsup{0.6}}$ & 81.3$_{\smallsup{0.6}}$ & 83.1$_{\smallsup{0.2}}$ & 89.9$_{\smallsup{0.2}}$ \\
\ssc{PromptTuning}  & 59.8$_{\smallsup{0.8}}$ & 63.1$_{\smallsup{1.1}}$ & 74.5$_{\smallsup{2.2}}$ & 79.2$_{\smallsup{0.9}}$ & 88.8$_{\smallsup{0.2}}$ \\
\ssc{Multi-taskModelTuning} & \textbf{64.6}$_{\smallsup{0.2}}$ & \textbf{79.2}$_{\smallsup{0.3}}$ & \textbf{84.5}$_{\smallsup{0.1}}$ & 88.0$_{\smallsup{0.5}}$ & 90.1$_{\smallsup{0.2}}$ \\
\ssc{\spot (Ours)} & 64.5$_{\smallsup{0.3}}$ & 73.2$_{\smallsup{0.3}}$ & 82.7$_{\smallsup{0.2}}$ & \textbf{88.7}$_{\smallsup{0.3}}$ & \textbf{91.2}$_{\smallsup{0.1}}$ \\
\midrule
\end{tabular}
\end{adjustbox}
\caption{\ssc{SuperGLUE} performance of different model tuning and prompt tuning methods across model sizes. We report the mean and standard deviation (in the subscript) across three random seeds. \spot outperforms vanilla %
\ssc{PromtTuning} and \ssc{\mbox{GPT-3}} by a large margin, matching or outperforming
\ssc{ModelTuning} across all model sizes. At the \ssc{XXL} model size, \spot even outperforms %
\ssc{Multi-taskModelTuning}, which fine-tunes the entire model on the \ssc{GLUE} mixture before fine-tuning it on individual \ssc{SuperGLUE} tasks.}
\label{tbl:fig1}
\end{table}

\vspace{2mm}
\begin{table*}[t!]
\centering
\begin{adjustbox}{max width=\textwidth}
\begin{tabular}{ c c c c c c c c c c c c c}
\toprule
\multirow{2}{*}{} & \multirow{2}{*}{\textbf{Model}} & \textbf{Total} & \textbf{Tuned} & \multirow{2}{*}{\bssc{Score}} & \multirow{2}{*}{\bssc{BoolQ}} & \multirow{2}{*}{\bssc{CB}} & \multirow{2}{*}{\bssc{COPA}} & \multirow{2}{*}{\bssc{MultiRC}} & \multirow{2}{*}{\bssc{ReCoRD}} & \multirow{2}{*}{\bssc{RTE}} & \multirow{2}{*}{\bssc{WiC}}	& \multirow{2}{*}{\bssc{WSC}} \\
& & \textbf{parameters} & \textbf{parameters} & & & & & & & & & \\ 
\midrule
\midrule
\multirow{5}{*}{\shortstack{\textbf{Top-\aclcr{7}} \\
\textbf{submissions}}} & \aclcr{\ssc{ST-MoE-32B}} & \aclcr{269B} & \aclcr{269B} & \aclcr{\textbf{91.2}} & \aclcr{92.4} & \aclcr{96.9/98.0} & \aclcr{99.2} & \aclcr{89.6/65.8} & \aclcr{95.1/94.4} & \aclcr{93.5} & \aclcr{77.7} & \aclcr{96.6} \\
& \aclcr{\ssc{Turing NLR v5}} & \aclcr{5.4B} & \aclcr{5.4B} & \aclcr{90.9} & \aclcr{92.0} & \aclcr{95.9/97.6} & \aclcr{98.2} & \aclcr{88.4/63.0} & \aclcr{96.4/95.9} & \aclcr{94.1} & \aclcr{77.1} & \aclcr{97.3} \\
& \ssc{ERNIE 3.0} & 12B & 12B & 90.6 & 91.0 & 98.6/99.2 & 97.4 & 88.6/63.2 & 94.7/94.2 & 92.6 & 77.4	 & 97.3 \\
& \ssc{T5 + UDG} & 11B & 11B & 90.4 & 91.4 & 95.8/97.6 & 98.0 & 88.3/63.0 & 94.2/93.5 & 93.0 & 77.9 & 96.6 \\
& \ssc{DeBERTa / TuringNLRv4} & 3.1B & 3.1B & 90.3 & 90.4 & 95.7/97.6 & 98.4 & 88.2/63.7 & 94.5/94.1 & 93.2 & 77.5 & 95.9 \\
& \ssc{Human Baselines} & - & - & 89.8 & 89.0 & 95.8/98.9 & 100.0 & 81.8/51.9 & 91.7/91.3 & 93.6 & 80.0 & 100.0 \\
& \ssc{T5} & 11B & 11B & 89.3 & 91.2 & 93.9/96.8 & 94.8 & 88.1/63.3 & 94.1/93.4 & 92.5 & 76.9 & 93.8 \\
\midrule
\midrule
\multirow{4}{*}{\shortstack{\textbf{Parameter-} \\ \textbf{efficient} \\ \textbf{adaptation} }} & \ssc{\textcolor{spotcolor}{Frozen T5 1.1 + SPoT}} & 11B & 410K & \textbf{89.2} & 91.1 & 95.8/97.6 & 95.6 & 87.9/61.9 & 93.3/92.4 & 92.9 & 75.8 & 93.8 \\
& \ssc{GPT-3 few-shot} & 175B & 0 & 71.8 & 76.4 & 52.0/75.6 & 92.0 & 75.4/30.5 & 91.1/90.2 & 69.0 & 49.4 & 80.1 \\
& \ssc{WARP few-shot} & 223M & 25K & 48.7 & 62.2 & 70.2/82.4 & 51.6 & 0.0/0.5 & 14.0/13.6 & 69.1 & 53.1 & 63.7 \\
& \ssc{CBoW} & 15M & 33K & 44.5 & 62.2 & 49.0/71.2 & 51.6 & 0.0/0.5 & 14.0/13.6 & 49.7 & 53.1 & 65.1 \\
\bottomrule
\end{tabular}
\end{adjustbox}
\caption{\ssc{SuperGLUE} results of our \spot{} \ssc{xxl} submission (in \textcolor{spotcolor}{green}) and competitors from the leaderboard as of \aclcr{2022/02/09}.}%
\label{tbl:superglue_submission}
\end{table*}

\section{Additional training details}
\label{appendix:training_details}

For \ssc{PromptTuning}, following~\citet{BLester21}, we initialize the prompt tokens with embeddings that represent an enumeration of the output classes with a back off to sampled vocabulary to fill any remaining prompt positions.

For model tuning approaches, we use the default hyperparameters for \ssc{T5}~\cite{CRaffel20}, i.e., learning rate 0.001,  Adafactor optimizer with pre-training parameter states restored, and dropout probability 0.1. 
To improve the model tuning baselines, we perform a sweep over the batch size hyperparameter and select $2^{16}$ tokens per batch, following~\citet{BLester21}. 

\section{Details of our \bssc{SuperGLUE} submission}
\label{appendix:superglue}
Table~\ref{tbl:superglue_submission} shows the performance of our \ssc{SPoT XXL} \ssc{SuperGLUE} submission, along with several strong competitors from the public \ssc{SuperGLUE} leaderboard. Apart from the human baseline, the top-\aclcr{7} submissions all tune \textgreater{}3B parameters directly on the final tasks. Only three previous \ssc{SuperGLUE} submissions use \emph{parameter efficient adaptation}, in the sense of tuning \textless{}1M parameters on the final tasks; all other submissions tune \textgreater{}50M parameters.\footnote{The ``AILabs Team, Transformers'' submission is listed as tuning 3M parameters, but we suspect this is in error, as the submission mentions using the \fssc{T5-3B} and \fssc{T5-Large} models.}

Our \spot submission achieves a score of \aclcr{89.2}, which far exceeds all other parameter-efficient adaptation methods, including \ssc{\mbox{GPT-3}}, which benefits from over 10$\times$ more frozen parameters (although it uses no tuned parameters). Compared to \ssc{WARP}~\cite{KHambardzumyan21}, our \spot approach tunes 16$\times$ more parameters (410K vs.~25K), and benefits from 50$\times$ more frozen parameters.

To the best of our knowledge, \spot is the first parameter-efficient adaptation approach that is competitive with methods that tune billions of parameters. Most notably, \spot{}'s performance \aclcr{almost} matches that of fully fine-tuned \ssc{T5 XXL} (89.3), despite building on the same underlying model, and tuning 27,000$\times$ fewer parameters. We note that \spot outperforms \ssc{T5} on three of eight \ssc{SuperGLUE} tasks (namely, \ssc{CB}, \ssc{COPA}, \ssc{RTE}). 

\section{Task transferability results}
\label{appendix:transfer_results}

The full results of our task transferability experiments can be found in Table~\ref{transfer_results}. We show that in many cases, initializing the prompt to that of a source task can provide significant gain on a target task. Table~\ref{effective_transfers} displays positive transfers with more than 10\% relative error reduction on the target task.

\section{Task embedding similarity}
\label{appendix:task_embedding_similarity}
\begin{figure*}[t!]
\centering
\includegraphics[width=\textwidth]{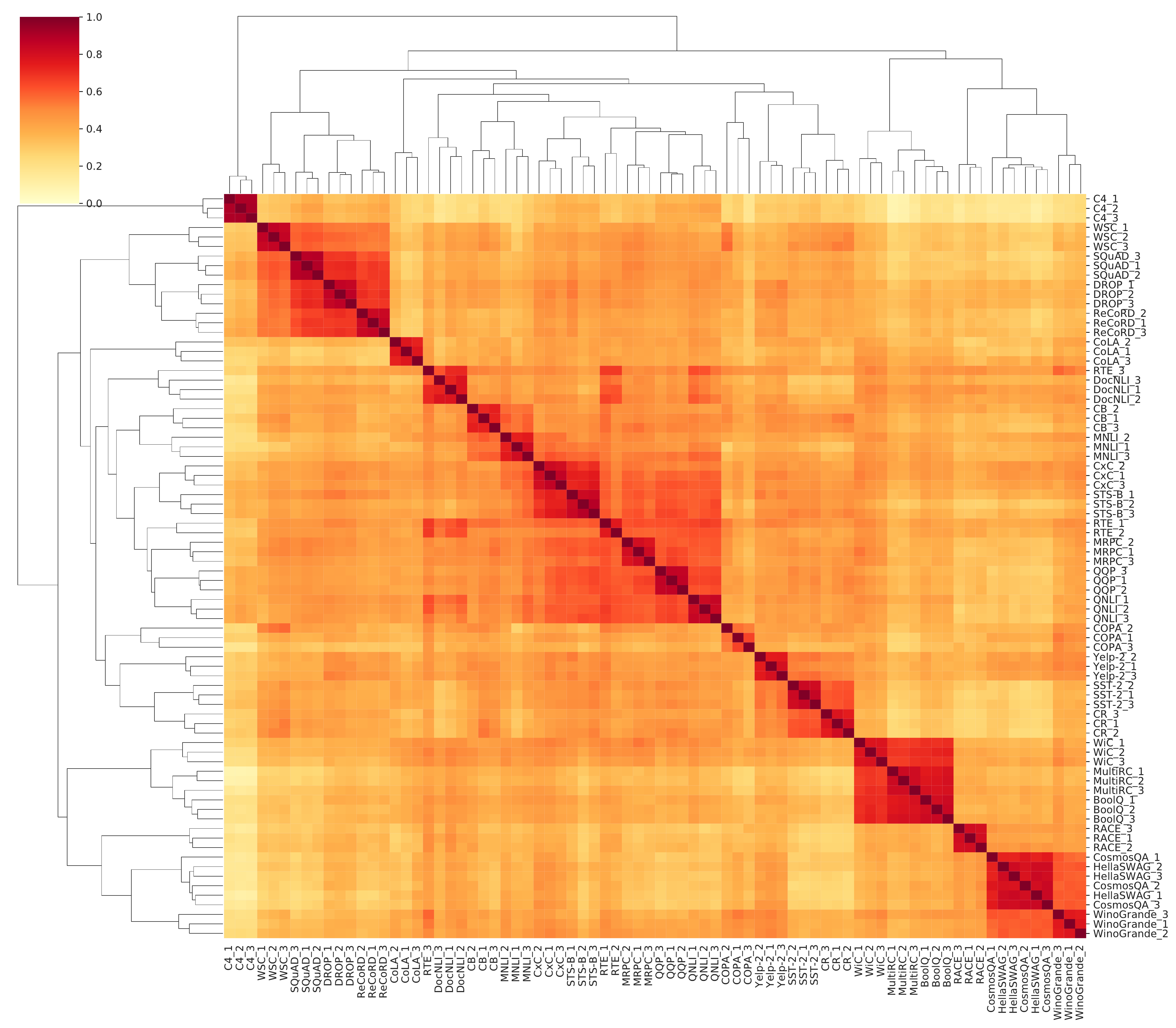}
\caption{%
Our prompt-based task embeddings capture task relationships: similar tasks group together into clusters. Additionally, task embeddings that are derived from different prompts of the same task are linked together. $t\_1, t\_2$, $t\_3$ correspond to three different prompt tuning runs on task $t$.}
\vspace{-2mm}
\label{fig:prompt_clustermap}
\end{figure*}

In Figure~\ref{fig:prompt_clustermap}, we show a clustered heatmap of cosine similarities between the task embeddings of the 26 \ssc{NLP} tasks we study in our task transferability experiments. For each task, we include the resulting task embeddings from all the three different prompt tuning runs on the task. As can be seen, our task embeddings capture task relationships: similar tasks cluster together. Additionally, task embeddings that are derived from different prompts of the same task are linked together.
\section{Correlation between task similarity and task transferability}
\begin{figure*}[t]
\centering
\includegraphics[width=\textwidth]{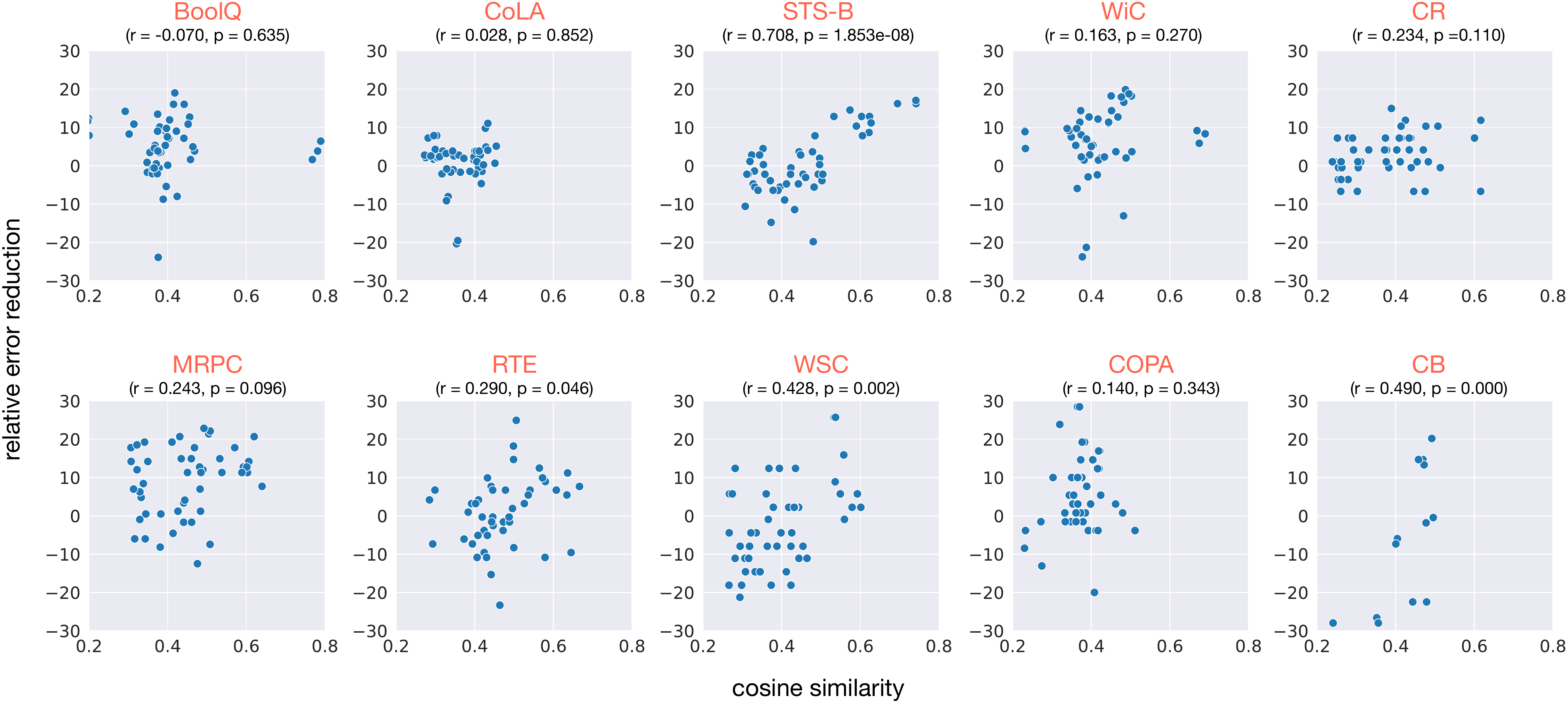}
\caption{Correlation between task similarity and task transferability. Each point represents a source prompt. The x-axis shows the cosine similarity between the associated source and target task embeddings, averaged over three runs for the target task (orange title). The y-axis measures the relative error reduction on the target task achieved by each source prompt. We include the Pearson correlation coefficient ($r$) and p-value.} %
\label{fig:similarity_vs_transferability_full}
\vspace{-2mm}
\end{figure*}

Figure~\ref{fig:similarity_vs_transferability_full} shows how the relative error reduction on a target task changes as a function of the similarity between the source and target task embeddings.
\label{appendix:correlation_similarity_transferability}
\newpage
\newpage
\begin{table*}[t!]
\centering
\begin{adjustbox}{max width=\textwidth}
\begin{tabular}{ l l l l l l l l l l l}
\toprule
& \bssc{BoolQ} & \bssc{CoLA} & \bssc{STS-B} & \bssc{WiC} & \bssc{CR} & \bssc{MRPC} & \bssc{RTE}  & \bssc{WSC} & \bssc{COPA} & \bssc{CB} \\
\bssc{Baseline} & 
\textcolor{myorange}{73.0}$_{1.2}$ &
\textcolor{myorange}{52.9$_{1.2}$} & \textcolor{myorange}{88.1}$_{0.6}$ &
\textcolor{myorange}{63.6}$_{1.6}$ &
\textcolor{myorange}{93.5}$_{0.2}$ & \textcolor{myorange}{86.1}$_{0.7}$ & \textcolor{myorange}{68.7}$_{1.2}$ &   \textcolor{myorange}{71.5}$_{1.7}$ & \textcolor{myorange}{56.7}$_{1.7}$ & \textcolor{myorange}{92.7}$_{1.9}$ \\
\bssc{C4} & \textcolor{mygreen}{75.8}$_{0.5}$ & \textcolor{mygreen}{54.8}$_{1.1}$ & \textcolor{black}{87.8}$_{0.6}$ & \textcolor{mygreen}{66.3}$_{0.8}$ & \textcolor{mygreen}{\textbf{93.9}}$_{0.1}$ & \textcolor{mygreen}{88.0}$_{0.6}$ & \textcolor{mygreen}{69.1}$_{1.9}$ & \textcolor{black}{68.0}$_{0.5}$ & \textcolor{black}{54.3}$_{0.9}$ & \textcolor{black}{83.1}$_{5.7}$  \\
\bssc{DocNLI} & \textcolor{black}{72.7}$_{1.4}$ & \textcolor{black}{52.7}$_{0.9}$ & \textcolor{black}{87.3}$_{0.9}$ & \textcolor{mygreen}{64.7}$_{0.3}$ & \textcolor{mygreen}{93.6}$_{0.4}$ & \textcolor{mygreen}{86.2}$_{0.8}$ & \textcolor{black}{67.4}$_{2.6}$ & \textcolor{black}{71.1}$_{3.6}$ & \textcolor{black}{56.0}$_{5.9}$ & \textcolor{black}{87.2}$_{1.7}$  \\
\bssc{Yelp-2} & \textcolor{mygreen}{74.8}$_{0.7}$ & \textcolor{mygreen}{53.9}$_{0.2}$ & \textcolor{black}{88.1}$_{0.3}$ & \textcolor{mygreen}{64.7}$_{0.5}$ & \textcolor{mygreen}{93.8}$_{0.3}$ & \textcolor{mygreen}{86.6}$_{0.8}$ & \textcolor{mygreen}{69.2}$_{1.1}$ & \textcolor{black}{70.8}$_{1.2}$ & \textcolor{black}{55.0}$_{0.0}$ & \textcolor{black}{87.8}$_{1.6}$  \\
\bssc{MNLI} & \textcolor{mygreen}{\textbf{77.6}}$_{0.4}$ & \textcolor{mygreen}{54.2}$_{0.7}$ & \textcolor{mygreen}{89.5}$_{0.3}$ & \textcolor{mygreen}{69.5}$_{0.5}$ & \textcolor{mygreen}{\textbf{93.9}}$_{0.4}$ & \textcolor{mygreen}{88.4}$_{0.6}$ & \textcolor{mygreen}{\textbf{74.7}}$_{1.3}$ & \textcolor{mygreen}{71.8}$_{3.3}$ & \textcolor{mygreen}{\textbf{69.3}}$_{2.1}$ & \textcolor{mygreen}{\textbf{97.0}}$_{1.1}$  \\
\bssc{QQP} & \textcolor{mygreen}{75.9}$_{0.5}$ & \textcolor{mygreen}{\textbf{55.6}}$_{1.3}$ & \textcolor{mygreen}{89.4}$_{0.2}$ & \textcolor{mygreen}{67.9}$_{0.2}$ & \textcolor{mygreen}{93.7}$_{0.5}$ & \textcolor{mygreen}{88.1}$_{0.7}$ & \textcolor{mygreen}{72.0}$_{0.5}$ & \textcolor{black}{71.5}$_{0.9}$ & \textcolor{mygreen}{62.0}$_{2.2}$ & \textcolor{black}{88.7}$_{4.2}$  \\
\bssc{QNLI} & \textcolor{mygreen}{75.6}$_{0.5}$ & \textcolor{mygreen}{55.5}$_{2.0}$ & \textcolor{mygreen}{89.2}$_{0.2}$ & \textcolor{mygreen}{69.6}$_{1.3}$ & \textcolor{mygreen}{93.8}$_{0.2}$ & \textcolor{mygreen}{87.8}$_{0.1}$ & \textcolor{mygreen}{71.1}$_{0.8}$ & \textcolor{black}{71.5}$_{2.5}$ & \textcolor{mygreen}{59.7}$_{3.9}$ & \textcolor{black}{92.5}$_{1.1}$  \\
\bssc{ReCoRD} & \textcolor{mygreen}{73.1}$_{0.9}$ & \textcolor{mygreen}{54.7}$_{1.3}$ & \textcolor{black}{87.7}$_{0.7}$ & \textcolor{mygreen}{65.5}$_{0.9}$ & \textcolor{mygreen}{93.7}$_{0.1}$ & \textcolor{mygreen}{\textbf{88.7}}$_{0.3}$ & \textcolor{black}{67.5}$_{1.3}$ & \textcolor{mygreen}{\textbf{77.2}}$_{2.3}$ & \textcolor{mygreen}{59.3}$_{1.2}$ & \textcolor{black}{74.1}$_{5.2}$  \\
\bssc{CxC} & \textcolor{mygreen}{75.9}$_{0.4}$ & \textcolor{mygreen}{55.0}$_{0.2}$ & \textcolor{mygreen}{\textbf{90.0}}$_{0.0}$ & \textcolor{mygreen}{\textbf{70.2}}$_{0.1}$ & \textcolor{mygreen}{\textbf{93.9}}$_{0.2}$ & \textcolor{mygreen}{88.0}$_{0.4}$ & \textcolor{mygreen}{70.3}$_{0.5}$ & \textcolor{black}{68.6}$_{2.5}$ & \textcolor{mygreen}{60.3}$_{3.9}$ & \textcolor{black}{89.3}$_{2.4}$  \\
\bssc{SQuAD} & \textcolor{mygreen}{76.0}$_{0.7}$ & \textcolor{mygreen}{54.9}$_{1.2}$ & \textcolor{black}{87.6}$_{0.1}$ & \textcolor{mygreen}{66.8}$_{0.3}$ & \textcolor{mygreen}{\textbf{93.9}}$_{0.5}$ & \textcolor{mygreen}{\textbf{88.7}}$_{0.7}$ & \textcolor{mygreen}{71.2}$_{0.4}$ & \textcolor{mygreen}{72.4}$_{0.5}$ & \textcolor{mygreen}{63.0}$_{1.6}$ & \textcolor{black}{91.3}$_{1.3}$  \\
\bssc{DROP} & \textcolor{mygreen}{73.6}$_{1.3}$ & \textcolor{mygreen}{53.0}$_{1.0}$ & \textcolor{black}{86.9}$_{0.9}$ & \textcolor{mygreen}{67.5}$_{1.2}$ & \textcolor{mygreen}{93.7}$_{0.2}$ & \textcolor{mygreen}{88.2}$_{0.3}$ & \textcolor{black}{65.7}$_{3.1}$ & \textcolor{mygreen}{73.4}$_{2.0}$ & \textcolor{mygreen}{60.0}$_{3.6}$ & \textcolor{black}{78.5}$_{8.6}$  \\
\bssc{SST-2} & \textcolor{mygreen}{73.3}$_{0.5}$ & \textcolor{black}{52.3}$_{0.3}$ & \textcolor{black}{87.9}$_{0.3}$ & \textcolor{mygreen}{63.8}$_{1.7}$ & \textcolor{mygreen}{93.8}$_{0.5}$ & \textcolor{black}{85.6}$_{0.9}$ & \textcolor{black}{66.9}$_{1.1}$ & \textcolor{black}{68.6}$_{0.4}$ & \textcolor{mygreen}{57.0}$_{2.2}$ & \textcolor{mygreen}{92.9}$_{1.3}$  \\
\bssc{WinoGrande} & \textcolor{mygreen}{74.1}$_{0.8}$ & \textcolor{black}{52.8}$_{1.6}$ & \textcolor{black}{87.8}$_{0.3}$ & \textcolor{black}{62.4}$_{2.5}$ & \textcolor{mygreen}{93.7}$_{0.1}$ & \textcolor{black}{86.1}$_{0.5}$ & \textcolor{black}{67.9}$_{1.3}$ & \textcolor{black}{71.5}$_{2.5}$ & \textcolor{black}{56.7}$_{1.2}$ & \textcolor{black}{83.9}$_{0.8}$  \\
\bssc{HellaSWAG} & \textcolor{black}{70.0}$_{2.6}$ & \textcolor{black}{32.7}$_{23.6}$ & \textcolor{black}{87.5}$_{0.2}$ & \textcolor{black}{60.1}$_{3.9}$ & \textcolor{mygreen}{93.6}$_{0.0}$ & \textcolor{mygreen}{86.6}$_{1.4}$ & \textcolor{black}{63.9}$_{5.4}$ & \textcolor{black}{70.2}$_{2.1}$ & \textcolor{mygreen}{58.0}$_{2.2}$ & \textcolor{black}{85.5}$_{2.6}$  \\
\bssc{MultiRC} & \textcolor{mygreen}{74.0}$_{0.5}$ & \textcolor{black}{50.0}$_{4.6}$ & \textcolor{mygreen}{88.2}$_{0.2}$ & \textcolor{mygreen}{66.4}$_{0.5}$ & \textcolor{black}{93.4}$_{0.1}$ & \textcolor{mygreen}{86.4}$_{1.3}$ & \textcolor{black}{67.6}$_{1.0}$ & \textcolor{black}{69.2}$_{4.1}$ & \textcolor{black}{56.0}$_{4.1}$ & \textcolor{black}{80.0}$_{8.6}$  \\
\bssc{CosmosQA} & \textcolor{mygreen}{73.4}$_{1.3}$ & \textcolor{black}{52.1}$_{2.3}$ & \textcolor{black}{87.7}$_{0.5}$ & \textcolor{mygreen}{65.9}$_{1.0}$ & \textcolor{mygreen}{93.6}$_{0.3}$ & \textcolor{mygreen}{87.9}$_{0.8}$ & \textcolor{black}{68.7}$_{1.6}$ & \textcolor{black}{69.6}$_{3.2}$ & \textcolor{mygreen}{62.3}$_{5.0}$ & \textcolor{black}{83.9}$_{8.8}$  \\
\bssc{RACE} & \textcolor{mygreen}{73.6}$_{0.5}$ & \textcolor{black}{52.5}$_{2.8}$ & \textcolor{black}{87.5}$_{0.5}$ & \textcolor{black}{63.1}$_{5.3}$ & \textcolor{black}{93.4}$_{0.2}$ & \textcolor{mygreen}{86.5}$_{0.8}$ & \textcolor{black}{66.5}$_{2.0}$ & \textcolor{black}{68.9}$_{1.2}$ & \textcolor{mygreen}{57.3}$_{1.2}$ & \textcolor{black}{84.8}$_{3.4}$  \\
\bottomrule
\end{tabular}
\end{adjustbox}
\caption{Many tasks can benefit each other via prompt transfer. The orange-colored row shows the results of prompt tuning \ssc{T5 Base} on the target tasks from scratch (i.e., without any prompt transfer). Each cell in the other rows represents the target task performance when transferring the prompt from the associated source task (row) to the associated target task (column). Positive transfers are shown in green and the best results are highlighted in bold (green). Numbers in the subscript indicate the standard deviation across 3 random seeds.}
\label{transfer_results}
\vspace{-2mm}
\end{table*}

\newpage
\begin{table*}[t!]
\centering
\footnotesize
\begin{adjustbox}{max width=0.4\textwidth}
\begin{tabular}{ l c }
\toprule
\textbf{Transfer} & \textbf{Increase (relative)} \\
& \\
\ssc{MNLI} $\rightarrow$ \ssc{CB} & 58.9 \\
\ssc{MNLI} $\rightarrow$ \ssc{COPA} & 29.1 \\  
\ssc{ReCoRD} $\rightarrow$ \ssc{WSC} & 20.0 \\  
\ssc{MNLI} $\rightarrow$ \ssc{RTE} & 19.2 \\
\ssc{ReCoRD} $\rightarrow$ \ssc{MRPC} & 18.7 \\  
\ssc{SQuAD} $\rightarrow$ \ssc{MRPC} & 18.7 \\
\ssc{CxC} $\rightarrow$ \ssc{WiC} & 18.1 \\
\ssc{MNLI} $\rightarrow$ \ssc{BoolQ} & 17.0 \\  
\ssc{MNLI} $\rightarrow$ \ssc{MRPC} & 16.5 \\ 
\ssc{QNLI} $\rightarrow$ \ssc{WiC} & 16.5 \\  
\ssc{MNLI} $\rightarrow$ \ssc{WiC} & 16.2 \\  
\ssc{CxC} $\rightarrow$ \ssc{STS-B} & 16.0 \\  
\ssc{DROP} $\rightarrow$ \ssc{MRPC} & 15.1 \\
\ssc{SQuAD} $\rightarrow$ \ssc{COPA} & 14.5 \\  
\ssc{QQP} $\rightarrow$ \ssc{MRPC} & 14.4 \\
\ssc{CxC} $\rightarrow$ \ssc{MRPC} & 13.7 \\  
\ssc{C4} $\rightarrow$ \ssc{MRPC} & 13.7 \\
\ssc{CosmosQA} $\rightarrow$ \ssc{MRPC} & 12.9 \\  
\ssc{CosmosQA} $\rightarrow$ \ssc{COPA} & 12.9 \\ 
\ssc{QQP} $\rightarrow$ \ssc{COPA} & 12.2 \\  
\ssc{QNLI} $\rightarrow$ \ssc{MRPC} & 12.2 \\  
\ssc{QQP} $\rightarrow$ \ssc{WiC} & 11.8 \\
\ssc{MNLI} $\rightarrow$ \ssc{STS-B} & 11.8 \\  
\ssc{SQuAD} $\rightarrow$ \ssc{BoolQ} & 11.1 \\  
\ssc{QQP} $\rightarrow$ \ssc{STS-B} & 10.9 \\
\ssc{QQP} $\rightarrow$ \ssc{BoolQ} & 10.7 \\ 
\ssc{CxC} $\rightarrow$ \ssc{BoolQ} & 10.7 \\ 
\ssc{DROP} $\rightarrow$ \ssc{WiC} & 10.7 \\
\ssc{QQP} $\rightarrow$ \ssc{RTE} & 10.5 \\
\ssc{C4} $\rightarrow$ \ssc{BoolQ} & 10.4 \\ 
\bottomrule
\end{tabular}
\end{adjustbox}
\caption{Positive transfers with more than 10\% relative error reduction on the target task. $s \rightarrow t$ denotes the transfer from source task $s$ to target task $t$.}
\label{effective_transfers}
\vspace{-2mm}
\end{table*}


\label{sec:appendix}

\end{document}